\documentclass[sigconf,acmart]{acmart}
\usepackage{xspace}
\usepackage{multirow} 
\PassOptionsToPackage{numbers, compress}{natbib}
\def\dataset{{\textsc{FinFraud-Real}}\xspace}
\def\rag{{\textsc{AuditAgent}}\xspace}
\newcommand{\nosection}[1]{\vspace{3pt}\noindent\textbf{#1.}}
\AtBeginDocument{%
  }

\setcopyright{acmlicensed}
\copyrightyear{2025}
\acmYear{2025}
\acmDOI{XXXXXXX.XXXXXXX}
\acmConference[ICAIF '25]{The Sixth ACM International Conference on AI in Finance}{November 15--18, 2025}{Singapore}
\acmISBN{978-1-4503-XXXX-X/2025/11}

\author{Songran Bai}
\affiliation{%
  \institution{Institute of Automation, Chinese Academy of Sciences}
  \city{Beijing}
  \country{China}
}
\email{songran.bai@mais.ia.ac.cn}

\author{Bingzhe Wu}
\affiliation{%
  \institution{Shenzhen University}
  \city{Shenzhen}
  \country{china}}
\email{wubingzheagent@gmail.com}

\author{Yiwei Zhang}
\affiliation{%
  \institution{Shenzhen University}
  \city{Shenzhen}
  \country{china}}
\email{2400671022@mails.szu.edu.cn}

\author{Chengke Wu}
\affiliation{%
  \institution{Shenzhen Institute of Advanced Technology, Chinese Academy of Sciences}
  \city{Shenzhen}
  \country{china}}
\email{ck.wu@siat.ac.cn}

\author{Xiaolong Zheng}
\affiliation{%
  \institution{Institute of Automation, Chinese Academy of Sciences}
  \city{Beijing}
  \country{China}
}
\email{xiaolong.zheng@ia.ac.cn}

\author{Yaze Yuan}
\affiliation{%
  \institution{Peking University}
  \city{Beijing}
  \country{China}
}
\email{yuanyaze@pku.org.cn}

\author{Ke Wu}
\affiliation{%
  \institution{Southern University of Science and Technology}
  \city{Shenzhen}
  \country{China}
}
\email{wuk@sustech.edu.cn}

\author{Jianqiang Li}
\affiliation{%
  \institution{Shenzhen University}
  \city{Shenzhen}
  \country{China}
}
\email{ljq@szu.edu.cn}

\begin{document}

%%
%% The "title" command has an optional parameter,
%% allowing the author to define a "short title" to be used in page headers.
\title{\rag: Expert-Guided Multi-Agent Reasoning for Cross-Document Fraudulent Evidence Discovery}

\begin{abstract}
Financial fraud detection in real-world scenarios presents significant challenges due to the subtlety and dispersion of evidence across complex, multi-year financial disclosures. In this work, we introduce a novel multi-agent reasoning framework \rag, enhanced with auditing domain expertise, for fine-grained evidence chain localization in financial fraud cases. Leveraging an expert-annotated dataset constructed from enforcement documents and financial reports released by the China Securities Regulatory Commission, our approach integrates subject-level risk priors, a hybrid retrieval strategy, and specialized agent modules to efficiently identify and aggregate cross-report evidence. Extensive experiments demonstrate that our method substantially outperforms General-Purpose Agent paradigm in both recall and interpretability, establishing a new benchmark for automated, transparent financial forensics. Our results highlight the value of domain-specific reasoning and dataset construction for advancing robust financial fraud detection in practical, real-world regulatory applications.
\end{abstract}

%%
%% The code below is generated by the tool at http://dl.acm.org/ccs.cfm.
%% Please copy and paste the code instead of the example below.
%%


%%
%% Keywords. The author(s) should pick words that accurately describe
%% the work being presented. Separate the keywords with commas.
%% A "teaser" image appears between the author and affiliation
%% information and the body of the document, and typically spans the
%% page.
% \begin{teaserfigure}
%   \includegraphics[width=\textwidth]{sampleteaser}
%   \caption{Seattle Mariners at Spring Training, 2010.}
%   \Description{Enjoying the baseball game from the third-base
%   seats. Ichiro Suzuki preparing to bat.}
%   \label{fig:teaser}
% \end{teaserfigure}

\begin{teaserfigure}
    \centering
    \includegraphics[width=\textwidth]{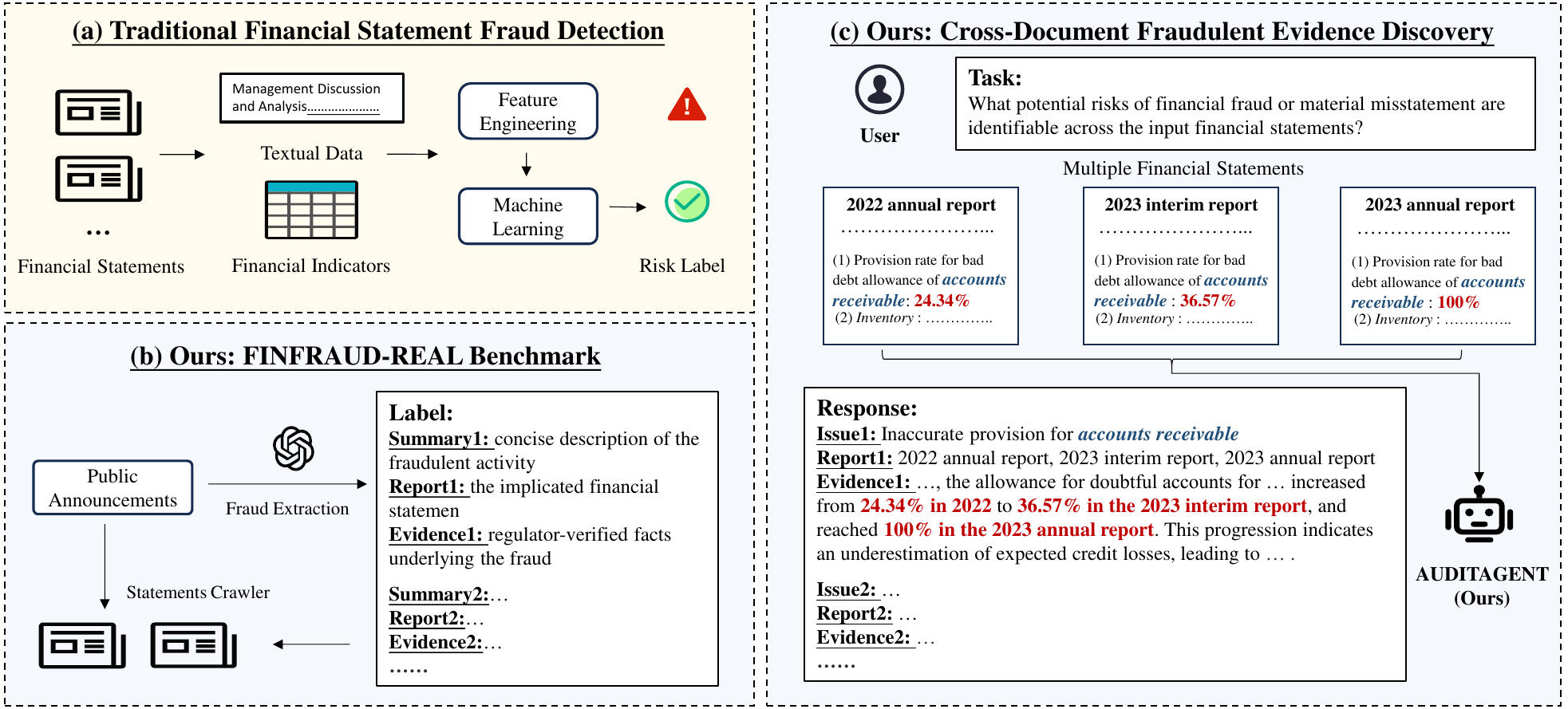}
    \caption{(a) Traditional fraud detection paradigm; (b) The construction process and data structure of \dataset; (c) The novel evidence identification task presented in this paper.}
    \Description{overall framework}
    \label{fig:task}
\end{teaserfigure}

% \received{20 February 2007}
% \received[revised]{12 March 2009}
% \received[accepted]{5 June 2009}

%%
%% This command processes the author and affiliation and title
%% information and builds the first part of the formatted document.
\maketitle

\section{Introduction}
\begin{quote}
\noindent
    \textit{This is something I've never mentioned before: I spend more time on balance sheets—unlike Wall Street, which pays them little attention—because, over time, they reveal more and are harder to manipulate than income statements. }\\
    --- Warren Buffett, 2025 Berkshire Hathaway Shareholders Meeting 
\end{quote}
Financial statement fraud has long been an issue of significant concern to regulatory authorities, investors, auditors, and stakeholders worldwide, causing profound economic losses, damaging investor confidence, and threatening market integrity~\cite{xin2018economic}.
Historically, fraud detection in financial statements has primarily relied on econometric and auditing methods grounded on predefined theoretical frameworks such as the Fraud Triangle Theory~\cite{sanchez2021fraud}, Fraud Diamond Model~\cite{umar2020fraud}, and Pentagon Theory~\cite{apriliana2017analysis}. These traditional approaches typically involve manually engineered financial indicators extracted from statements, such as financial ratios and macro-level accounting attributes~\cite{ashtiani2021intelligent}. Subsequently, machine learning (ML) algorithms are developed based on these extracted features to predict fraudulent activities~\cite{ali2022financial,campa2023roles}. Despite some promising results, these approaches fail to elucidate the underlying reasons for fraud clearly, nor can they pinpoint the exact financial statement segments serving as evidence of fraudulent reporting~\cite{nesvijevskaia2021accuracy}. This traditional workflow is conceptually shown in Figure \ref{tab:overall} (a). Consequently, their practical applicability in auditing and regulatory scenarios remains significantly constrained.

Recent advancements in Large Language Models (LLMs), such as GPT-4 and DeepSeek R1~\cite{zhao2023survey}, have demonstrated remarkable capabilities in natural language understanding~\cite{karanikolas2023large}, reasoning~\cite{chen2025towards}, and context-aware analysis~\cite{liu2025comprehensive}. Inspired by these powerful capabilities, this paper introduces a more challenging and practically meaningful task compared to previous discriminative prediction paradigm as shown in Figure \ref{tab:overall} (c): accurately retrieving explicit textual evidence of fraudulent activities across financial statements spanning multiple fiscal years. This fine-grained evidence identification not only enhances prediction accuracy but also significantly improves interpretability by clearly elucidating the reasoning paths associated with fraudulent statements.

However, exploiting LLMs for this task poses two critical challenges:
First, the lack of realistic benchmarks significantly hinders progress. Existing financial domain benchmarks primarily target relatively straightforward numerical reasoning tasks~\cite{shen2023positional} and factual question-answering~\cite{zhang2023survey}, which inadequately reflect the complexity and subtlety inherent in real-world financial fraud scenarios~\cite{hilal2022financial}. 
To address this gap, this study makes a substantial contribution in dataset construction by introducing a novel benchmark named \dataset as shown in Figure \ref{tab:overall} (b). Leveraging publicly disclosed enforcement documents and financial reports from the China Securities Regulatory Commission~\cite{csrcCSRC}, we systematically curate and annotate a real-world dataset comprising multi-year, multi-document financial statements and detailed records of confirmed fraudulent behaviors. Each case is enriched with expert-verified issues, capturing both the implicated accounting subjects and the supporting textual evidence. This resource enables rigorous evaluation of evidence chain localization methods and provides the research community with a high-quality benchmark for advancing automated financial fraud detection.

Second, fraudulent activities are typically concealed and sophisticated, with evidence often dispersed subtly across financial disclosures spanning multiple fiscal years~\cite{reurink2019financial}. General-Purpose Agent paradigms—such as ReAct~\cite{yao2023react}, Search-o1~\cite{li2025search} and related frameworks~\cite{wang2024survey} are designed for broad task applicability, but often lack the domain-specific reasoning strategies and deep contextual awareness required to uncover intricate, cross-year fraud patterns in financial data. These approaches may struggle to recognize subtle, contextually dispersed indicators of fraud that require both expert knowledge and multi-document synthesis. To address these challenges, we introduce a novel domain-specific agent framework named \rag, which embeds auditing expertise directly into each stage of our framework, designing specialized agent modules and retrieval strategies tailored for the unique demands of financial fraud detection. Specifically, \rag first constructs a variational bayesian-based model to learn subject-level fraud risk priors from historical enforcement data provided by the China Securities Regulatory Commission. These priors guide a hybrid sparse-dense retrieval strategy, enabling efficient identification of relevant evidence across multiple, lengthy financial reports. Building upon the retrieved evidence, we design a multi-expert reasoning framework based on LLMs such as DeepSeek R1~\cite{guo2025deepseek}, in which specialized reasoning models perform intra-document and cross-document risk analysis, followed by global evidence aggregation. 

Extensive experiments on our curated dataset demonstrate that the proposed approach achieves significant improvements in both the recall and interpretability of financial fraud evidence localization, outperforming General-Purpose Agent baselines by a substantial margin. Our results highlight the effectiveness of integrating domain expertise with advanced multi-agent reasoning, paving the way for more robust and transparent automated financial regulation in real-world regulatory settings.

\section{Related Work}
\subsection{Machine Learning for Financial Statement Fraud Detection}
% Recent studies on financial statement fraud detection have primarily focused on extracting both textual features (e.g., MD\&A disclosures) and numerical ratios (e.g., debt-to-equity, cash flow margins) from financial statements, which are then fed into machine learning classifiers such as logistic regression, random forests, or neural networks for binary fraud prediction []. While these approaches demonstrate competitive accuracy, two critical limitations persist. First, widely adopted frameworks like the Beneish M-Score [], though effective in aggregating financial indicators, suffer from interpretability constraints due to their black-box formulations. Second, most data-driven methods [] rely on imbalanced binary labels (fraud/non-fraud), oversimplifying the heterogeneous nature of fraudulent practices. These opacity issues hinder practical adoption in auditing workflows, where regulators and stakeholders require transparent explanations—such as identifying specific anomalous transactions or ratio thresholds—to validate automated alerts.
Recent studies on financial statement fraud detection primarily focus on extracting textual content~\cite{DSS_Text_FinFraud} (e.g., MD\&A disclosures) and numerical indicators~\cite{DSS_metric_FinFraud} (e.g., revenue, expense ratios) from financial statements and applying machine learning models such as random forests or deep neural networks for binary classification tasks~\cite{ACM_ML_FinFraud, DSS_multimodal_FinFraud}. While achieving notable performance, these approaches exhibit two fundamental limitations. First, their supervisory signals typically depend on oversimplified binary labeling schemas (fraudulent and non-fraudulent), failing to capture the spectrum of tactic heterogeneity~\cite{IJAIS_binary_FinFraud}. Second, despite the empirical success of established frameworks, their interpretability remains constrained by the inherent opacity of their formulaic construction~\cite{ESWA_interpretability_FinFraud}. This explanatory deficit poses significant adoption barriers in regulatory auditing contexts, where audit trail transparency and decision traceability constitute essential requirements for both oversight agencies and corporate stakeholders.
% , making it challenging to align predictions with audit trails or regulatory requirements

\subsection{Financial Domain Benchmarks}
Recent progress in assessing LLMs for financial applications has led to the development of benchmarks focused on numerical reasoning, domain-specific terminology comprehension, and quantitative analytical tasks. For example, FinQA~\cite{chen2021finqa} and TAT-QA~\cite{zhu2021tat} focus on hybrid question answering over financial reports, requiring multi-step reasoning with textual and tabular data. ConvFinQA~\cite{chen2022convfinqa} extends this to conversational formats, modeling sequential reasoning in analyst workflows. Meanwhile, FinanceMATH~\cite{zhao2023financemath} and BizBench~\cite{koncel2023bizbench} emphasize knowledge-intensive math tasks and program synthesis for financial formulas, while FinBen~\cite{xie2024finben} and FinEval~\cite{zhang2023fineval} provide holistic frameworks covering diverse tasks such as financial sentiment and security analysis. However, these benchmarks primarily address isolated numerical reasoning or terminology tasks. They fall short in evaluating complex real-world scenarios such as cross-document evidence integration, temporal analysis, and identifying risk paths, which are critical challenges in applications like financial statement fraud detection. 

\subsection{LLMs-based Long Document Understanding}
Recent advancements in long document understanding have focused on enhancing the capabilities of LLMs through specialized frameworks. Retrieval-Augmented Generation (RAG) frameworks have evolved to mitigate the "lost in the middle" challenge in long-context QA~\cite{liu-etal-2024-lost}. LongRAG proposes a dual-perspective paradigm that preserves global document coherence while refining local factual retrieval, demonstrating robustness in multi-hop reasoning tasks~\cite{zhao-etal-2024-longrag}. Meanwhile, PEARL enhances action-oriented reasoning by decomposing queries into interpretable steps and executing them sequentially over long narratives~\cite{sun-etal-2024-pearl}. Several approaches also adopt multi-agent collaboration to address computational constraints and error propagation in lengthy document analysis. For instance, LongAgent employs a divide-and-conquer strategy, distributing document chunks across specialized agents for localized reasoning while leveraging a leader agent to synthesize outputs through iterative discussion~\cite{zhao-etal-2024-longagent}. Similarly, Chain-of-Agents (CoA) introduces hierarchical agent roles, where worker agents process segmented text and a manager agent consolidates results, enabling efficient information aggregation across extended contexts~\cite{NEURIPS2024_CoA}. However, these approaches are largely tailored for general domains and struggle with the precise, domain-specific demands of financial fraud detection.

\section{Data and Task: \dataset}

\nosection{Dataset} 
 To construct a dataset that reflects real-world financial fraud scenarios, we collected $N=1,570$ fraud cases disclosed by the China Securities Regulatory Commission (CSRC) over the past decade, specifically designed to support fine-grained evaluation of evidence chain localization in financial fraud detection. Formally, the dataset comprises $N$ independent samples $\mathcal{S} = \{(D_i, F_i)\}_{i=1}^N$, where for each sample $i$, $D_i = \{d_{i}^{1}, \ldots, d_{i}^{m_{i}}\}$ denotes a collection of $m_i$ associated financial statement documents containing of both unstructured narrative text and structured accounting tables. And $F_i$ represents the groud truth of fraud description provided by CSRC. 

% perform structured prediction based on the input document set $D_i$, aiming to recover a structured representation of the underlying fraud pattern.
\nosection{Structured Extraction of Fraudulent Behavior}
For each case, unstructured case descriptions released by the CSRC were processed using prompt engineering to guide LLMs in extracting three key dimensions: (i) \textit{Report}, denoting the implicated financial statements; (ii) \textit{Summary}, a concise description of the fraudulent activity; and (iii) \textit{Evidence}, regulator-verified facts underlying the fraud. To ensure reliable extraction, a validation team of three experts performed double-blind cross-validation on a random sample comprising 10\% of the cases. Finally, each sample $i$ is annotated with a set of $q_i$ fraudulent issues $F_i = \{(R_i^{j}, s_i^{j}, e_i^{j})\}_{j=1}^{q_i}$, where $R_i^{j} \subseteq D_i$ indicates the set of documents involved in the $j$-th fraud issue, $s_i^{j}$ provides a concise textual summary of the $j$-th fraudulent activity and $e_i^{j}$ contains the official detailed fact provided by the CSRC for $j$-th issue.
% 问题：标签中的evidence和从财报中得到的evidence不是在一个语义空间中

\nosection{Acquisition and Parsing of Original Financial Statements}
For extracted \textit{Report}, we developed an automated crawling system to obtain the corresponding financial statement PDFs from official disclosure platforms \cite{cninfox5DE8x6F6Ex8D44x8BAFx7F51}. Extracted PDFs were processed leveraging the PyMuPDF text extraction module \cite{githubGitHubPymupdfPyMuPDF}, alongside parsing strategies guided by standardized reporting formats and domain-specific knowledge. We extracted content from key sections including the main financial statements, management discussion and analysis, and notes to the consolidated financial statements. Furthermore, for reports identified as fraudulent, we augmented the dataset with financial statements from the two preceding reporting periods. Specifically, if fraud was detected in the 2022 annual report, we additionally retrieved the 2022 semi-annual report and the 2021 annual report to ensure comprehensive data coverage and integrity. 
% Ultimately, XX\% of all financial statements were successfully converted into structured textual data, with missing content supplemented as needed to maintain overall data integrity.

\nosection{Task} Unlike most prior studies that focus solely on binary classification—indicating the presence or absence of financial fraud—we introduce a more challenging and practically relevant task: precise matching of fraudulent activities within financial statements. The goal of the task is to generate a structured description of the underlying fraud pattern based on the input document set $D_i$. Specifically, the model must produce a set of $\hat{q_{i}}$  inferred fraudulent issues $\hat{F}_i = \{(\hat{R}_i^{j}, \hat{s}_{i}^{j}, \hat{E}_i^{j})\}_{j=1}^{\hat{q}_i}$, where $\hat{R}_i^{j} \subseteq D_i$ denotes the subset of documents implicated in the $j$-th inferred issue $\hat{s}_{j}$, $\hat{E}_i^{j}=\{\hat{e}_{i}^{jk}\}_{k=1}^{K}$ represents $K$ supporting evidence extracted from the documents, corresponding to the fraudulent behavior described in $\hat{s}_i^{j}$.
% which can be regarded as an open-ended Text Generation tasks

\section{Methodology: \rag}

\begin{figure*}
    \centering
    \includegraphics[width=0.9\textwidth]{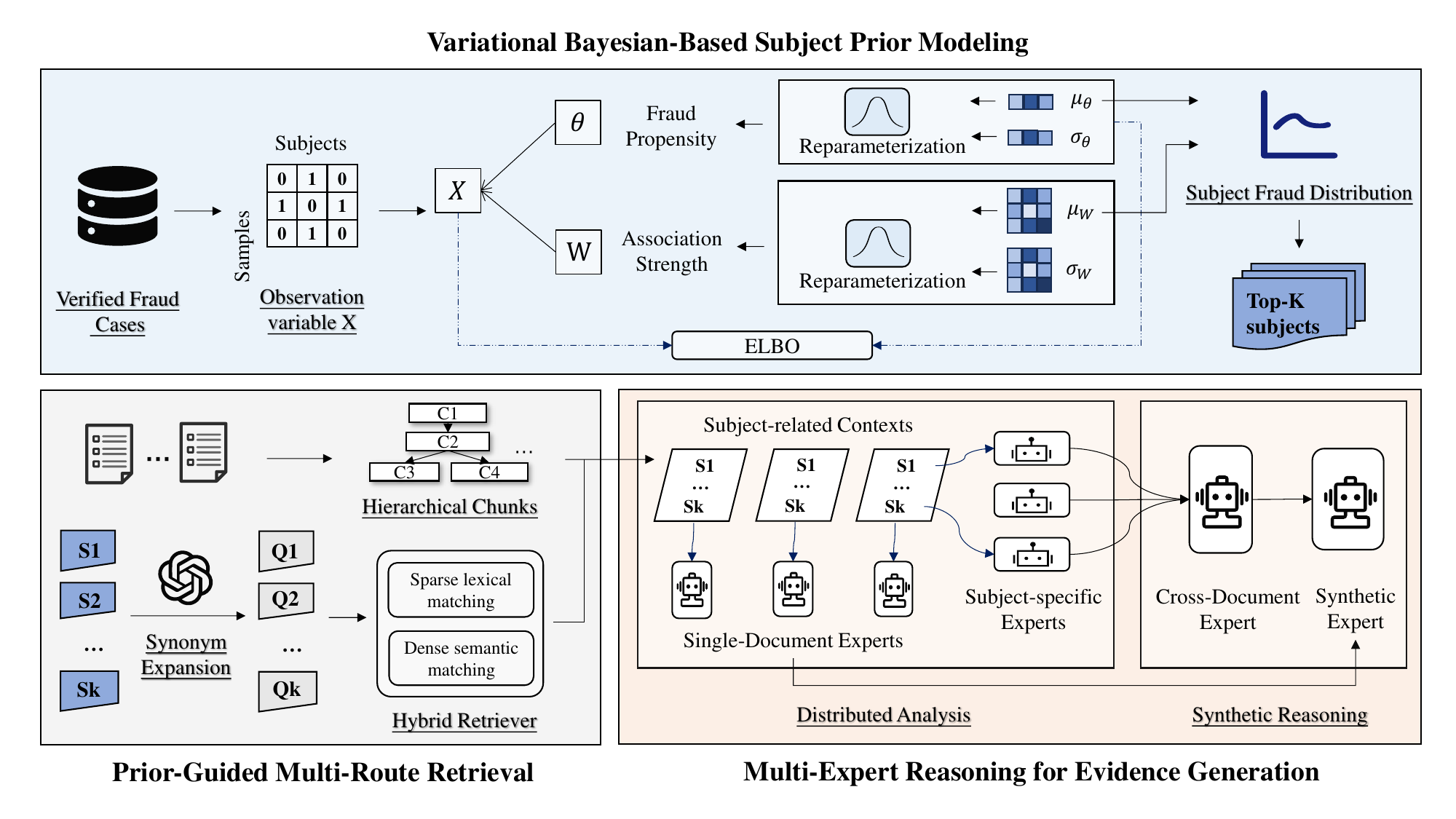}
    \caption{\rag consists of three key stages --- (1) Variational Bayesian-Based Subject Prior Modeling; (2) Prior-Guided Multi-Route Retrieval; (3) Multi-Expert Reasoning for Evidence Generation.}
    \label{fig:framework}
\end{figure*}

% We propose a novel evidence chain localization task requiring coordinated comprehension and analysis across multiple financial documents (averaging 43k tokens per case). Existing LLMs struggle in this setting due to limited context windows and weak support for long-range, multi-hop reasoning, while traditional single-agent frameworks often fail to retrieve relevant information or maintain sufficient memory. To address these limitations, 
We introduce \rag framework shown in Figure \ref{fig:framework} that emulates the cognitive and procedural expertise of human auditors throughout the analytical workflow. \rag integrates three key components: (1) a prior model of fraudulent behavior learning from real-world fraud cases, (2) a hybrid retrieval mechanism combining multiple strategies to improve coverage and precision, and (3) a dual-level auditing process that emulates expert strategies for both single-document and cross-document examination.

% Together, these components enable the agent to achieve fine-grained, interpretable localization of fraudulent evidence, even in complex, multi-document settings that challenge conventional LLMs and agent architectures.

\subsection{Variational Bayesian-Based Subject Prior Modeling}
\textbf{Motivation:} 
Detecting covert fraud evidence in extensive financial statements poses a needle-in-a-haystack challenge~\cite{NEURIPS2024_c0d62e70}. Inspired by accounting practices that strategically prioritize suspicious indicators (e.g., abnormal metric fluctuations) and analyze cross-subject coordination patterns~\cite{brazel2015understanding,elsayed2017indicators}, we propose an external fraud prior framework to inject domain knowledge into LLMs. Our approach achieves two objectives: (1) identifying high-risk subjects through decomposing fraud probability to intrinsic propensity and collaborative tendency, and (2) systematically narrowing evidence search scope via prioritized examination, enabling efficient localization of fraudulent evidence chains.

Our proposed prior modeling approach is grounded in the probabilistic graphical model framework, which treats subject-specific fraud propensity parameters and inter-subject association strength parameters as latent variables. Specifically, we define a binary observation matrix $X \in \{0,1\}^{N \times H}$, where $N$ denotes the number of fraud cases and $H$ epresents the total number of subjects. The entry $X_{ih}=1$ indicates that subject $h$ is involved in fraud case $i$. The latent variables include a vector $\theta \in \mathbb{R}^{H}$ and a symmetric association matrix $W \in \mathbb{R}^{H*H}$, where $\theta_{h}$ quantifies the base fraud propensity of subject $h$, and $W_{mn}$ captures the association strength between subject $m$ and $n$. Under the pseudo-likelihood assumption~\cite{SIGIR_pseudolikelihood}, the generative process for observed data can be expressed as:

\begin{equation}
p(X|\theta,W) \propto \prod_{i=1}^N \prod_{h=1}^H p(X_{ih})^{X_{ih}} \left(1-p(X_{ih})\right)^{1-X_{ih}}
\end{equation}

% $p(X_{ih})=\sigma\left(\theta_h + \sum_{j\neq h} W_{hj}X_{ij}\right)$ 
where $\sigma(\cdot)$ denotes the sigmoid function and $p(X_{ih}) = \sigma\Bigl( \theta_h + \sum_{j\neq h} W_{hj}X_{ij} \Bigr)$ represents the fraud probability. This formulation assumes the observed data $X$ is jointly generated by the latent variables $\theta$ and interaction matrix $W$. We further assume independent Gaussian priors for both $\theta \sim \mathcal{N}(0,\alpha^{-1}I)$ and $W \sim \mathcal{N}(0,\beta^{-1}I)$, with hyperparameters $\alpha$ and $\beta$ controlling the prior strength. To approximate the intractable posterior distribution $P(\theta,W|X)$, we employ variational inference by defining a factorized variational distribution $q_{\phi}(\theta, W)=q(\theta)q(W)$, where $q_{\theta}=\mathcal{N}(\mu_{\theta}, \sigma_{\theta})$ and $q_{W}=\mathcal{N}(\mu_{W}, \sigma_{W})$. The optimization objective is formulated as maximizing the evidence lower bound (ELBO)~\cite{WWW_Elbo}: 

\begin{equation}
    \mathcal{L} = \mathbb{E}_{q}[\log p(X|\tilde{\theta},\tilde{W})] - \text{KL}(q(\theta)|p(\theta)) - \text{KL}(q(W)|p(W))
\end{equation}

Through iterative maximization of the ELBO with synchronized updates of variational parameters, we obtain posterior estimates $\mu_{\theta}$ for subject-specific fraud propensities and $\mu_{W}$ for inter-subject associations. These estimates respectively characterize the inherent risk levels of individual subjects and cross-subject fraud transmission effects in financial misconduct scenarios.

% Specifically, we constructed a Markov Random Field (MRF) framework that leverages historical fraud cases to jointly infer both the marginal fraud probabilities of individual accounting subjects and the pairwise association strengths between them. Within this framework, each node represents an accounting subject, with node-level potential parameters $\theta_{\text{single}}$ quantifying a subject's intrinsic propensity for fraudulent manipulation. Undirected edges connect subjects that have co-occurred in historical fraud cases, where edge-level potential parameters $\theta_{\text{edge}}$ characterize the strength of collaborative fraud associations. Given labeled historical fraud event data, we implemented approximate inference through maximum likelihood estimation coupled with Gibbs sampling, enabling simultaneous learning of both $\theta_{\text{single}}$ and $\theta_{\text{edge}}$ parameters across the subject network. The final influence score for each subject was formulated as a linear combination of its nodal potential and associated edge potentials, allowing the model to prioritize high-risk subjects during downstream audit retrieval and evidential reasoning tasks. This expert-informed prior modeling establishes a principled framework for attention guidance in large-scale, multi-document financial fraud detection tasks.

\subsection{Prior-Guided Multi-Route Retrieval}

Given a financial statement $d$, we first decompose it into semantically coherent chunks $\mathcal{C} = \{c_i\}_{i=1}^M$ through structure parsing of their inherent logical hierarchy, where each $c_i$ corresponds to a subsection identified through document organization patterns (e.g., ``Notes to Consolidated Financial Statements'' → ``Accounts Receivable''). For $K$ high-risk subjects $\mathcal{S} = \{s_k\}_{k=1}^K$ identified via prior importance distribution  $\mathcal{P}(s)$, we generate domain-adapted query variants $Q_k$ via LLMs-driven synonym expansion, where the expansion incorporates domain-specific mappings (e.g., ``Inventory'' $\rightarrow$ \{``Inventory'', ``Provision for inventory depreciation'', ``Inventory impairment loss''\}). The retrieval process employs a hybrid architecture with parallelized subject-specific query execution, combining sparse lexical mathing and dense semantic matching through neural embeddings with cosine distance. This multi-route design operates concurrently across all $|\mathcal{S}|$ subjects, prioritizing content blocks annotated with subject-specific subsection---``Notes to Consolidated Financial Statements'' identified through hierarchical parsing. The resultant evidentiary corpus forms the foundation for subsequent contextual reasoning and evidentiary chain construction.

\subsection{Multi-Expert Reasoning for Evidence Generation}
\textbf{Motivation:} 
Detecting financial fraud requires identifying subtle patterns in long-range document dependencies through multi-hop reasoning (e.g., mismatched cash flow and revenue trends). Current methods fail to concurrently capture intra-document contradictions and inter-document temporal anomalies while preserving audit-interpretable evidence chains. We propose a multi-expert LLM framework integrating distributed analysis with hierarchical evidence fusion to address these challenges.
% v2
% Financial statements fraud detection requires identifying subtle patterns embedded in long-range contextual dependencies across documents, which often demand multi-hop reasoning. For instance, a company might exhibit anomalous operating cash flow patterns characterized by mismatched trends between cash flow and revenue growth. Existing approaches struggle to simultaneously capture intra-document contradictions and inter-document temporal anomalies while maintaining audit-interpretable evidence chains. This motivates our design of a collaborative LLMs-based multi-expert framework that combines distributed analysis with hierarchical evidence integration.
% v1
% Building upon prior high-risk accounting subjects and their associated contexts, we aim to aggregate these elements and infer potential fraudulent issues along with corresponding evidence chains. However, fraudulent patterns are frequently embedded in long-range contextual dependencies and require complex multi-hop reasoning processes. For instance, a company might exhibit anomalous operating cash flow patterns characterized by mismatched trends between cash flow and revenue growth. Detecting such manipulation necessitates cross-document navigation to locate relevant financial metrics, followed by multi-step logical validation through trend analysis and ratio evaluation to establish evidentiary chains. To achieve this granular and comprehensive risk analysis, we developed a multi-expert framework that collaboratively executes both intra-report and inter-report risk reasoning.
The framework operates through two phases with coordinated expert interactions: 

\textbf{Phase 1: Distributed Analysis} 
The $T$ single-document experts apply domain-specific financial logic and accounting principles to detect contradictions and risk signals $\{O_{s}^{(t)}\}_{t=1}^{T}$ within high-risk subjects in individual document. Concurrently, the $K$ subject-specific experts analyze periodic trend variations $\{O_{t}^{(k)}\}_{k=1}^{K}$ of accounting metrics associated with these high-risk subjects. 

\textbf{Phase 2: Synthetic Reasoning} 
These temporal patterns are aggregated by a cross-document expert through the operation $O_{c} = \mathrm{LLM}(\{O_{t}^{(k)}\}_{k=1}^{K})$, identifying suspicious cross-subject correlations across multiple time periods. The synthetic expert then integrates evidence from both single-document analyses and cross-document observations to construct a holistic risk assessment via $O = \mathrm{LLM}(\{O_{s}^{(t)}\}_{t=1}^{T}, O_{c})$. This meta-expert performs three key functions: resolving overlapping alerts through probabilistic conflict resolution, reconciling contradictory signals via hierarchical reasoning, and constructing interpretable evidentiary chains to support audit procedures. The multi-expert architecture emulates real-world audit team dynamics through coordinated agent communication, enabling long-range reasoning capabilities for robust detection of complex fraud patterns.

\section{Experiments}
\subsection{Experimental Setup}
% Our experiments are designed to evaluate two key aspects of \rag. First, we assess the effectiveness of incorporating fraud risk priors and domain expert knowledge into the multi-agent reasoning architecture. Compared to General-Purpose Agent framework and conventional single-document LLMs reasoning baselines, our method achieves a substantial improvement in accurately matching real-world fraud risks and localizing fine-grained evidence. Second, we provide an analysis of the prior distribution of fraudulent behaviors, extracted from a large corpus of historical enforcement cases. This analysis not only informs the design of our risk-guided retrieval strategy, but also offers valuable insights into the patterns and prevalence of specific fraud schemes in the financial domain. 

% \nosection{Baseline} To rigorously evaluate the effectiveness of \rag, we compare it against two major categories of baselines: (1) Single LLMs, where the input is truncated to ensure reasoning completeness within the context window; and (2) General-Purpose Agent framework such as Search-o1\cite{li2025search}, in which the model autonomously determines strategies for searching and assembling chains of fraudulent evidence. For a comprehensive comparison, we employ both General LLMs, Domain LLMs and Reasoning-Augmented LLMs as backbone, including Qwen2.5-32B~\cite{yang2024qwen2}, DeepSeek-V3~\cite{liu2024deepseek}, DianJin-R1-32B~\cite{zhu2025dianjin}, QwQ 32B~\cite{qwenlmQwQReflect} and Deepseek R1~\cite{guo2025deepseek}.
\nosection{Baseline} To rigorously evaluate the effectiveness of \rag, we compare it against two major categories of baselines: (1) Single LLMs, where the input is truncated to ensure reasoning completeness within the context window; and (2) General-Purpose Agent framework, such as Search-o1~\cite{li2025search}, where the model autonomously determines strategies for retrieving and assembling chains of evidence, including potentially deceptive information. For a comprehensive comparison, we employ both General LLMs, Domain LLMs and Reasoning-Augmented LLMs as backbone. Specifically, the baselines include Qwen2.5-32B~\cite{yang2024qwen2}, DeepSeek-V3~\cite{liu2024deepseek}, DianJin-R1-32B~\cite{zhu2025dianjin}, QwQ-32B~\cite{qwenlmQwQReflect} and DeepSeek-R1~\cite{guo2025deepseek}.

\nosection{Metric} Given that this study focuses on detecting deeply concealed fraudulent activities—rather than maximizing overall accuracy—we prioritize recall over precision at both levels. This emphasis reflects real-world auditing practices, where experts often cannot comprehensively identify all potential risks. An automated system that highlights a broad set of suspicious cases, even with some false positives, can serve as an effective early-warning mechanism and aid auditors in uncovering hidden fraud. Furthermore, since a single case may involve multiple fraud types and LLM-based systems can generate multiple risk assessments per instance, we propose a customized \textit{many-to-many} recall metric to better capture model performance in this complex, multi-output setting.

\begin{table*}
\caption{Overall results of \rag.}
\label{tab:overall}
\centering
% \footnotesize
\setlength{\tabcolsep}{3pt} % 调整列间距
% \fontsize{8}{9.6}\selectfont
\begin{tabular}{lccccccc}
\toprule
& & \multicolumn{2}{c}{Single LLMs} & \multicolumn{2}{c}{General-Purpose Agent} & \multicolumn{2}{c}{\textbf{\rag (Ours)}} \\
\cmidrule(lr){3-4} \cmidrule(lr){5-6} \cmidrule(lr){7-8}
Category & Model & $R_{I}$ & $R_{E}$ & $R_{I}$ & $R_{E}$ & $R_{I}$ & $R_{E}$ \\
\midrule
\multirow{2}{*}{General LLMs} 
& Qwen2.5-32B    & 17.52 & 9.52 & 8.11 & 5.47 & \textbf{19.68} & \textbf{12.78} \\
& DeepSeek-V3    & 24.71 & 17.99 & 13.25 & 10.28 & \textbf{26.62} & \textbf{21.34} \\
\midrule
Domain LLMs 
& DianJin-R1-32B & 21.94 & 17.84 & 11.17 & 10.69 & \textbf{23.87} & \textbf{18.87} \\
\midrule
\multirow{2}{*}{Reasoning-Augmented LLMs} 
& QwQ-32B        & 23.63 & 18.80 & 12.42 & 11.03 & \textbf{26.46} & \textbf{22.15} \\
& DeepSeek-R1    & 26.20 & 20.72 & 16.53 & 12.74 & \textbf{28.91} & \textbf{23.29} \\
\bottomrule
\end{tabular}
\end{table*}

% \begin{table}[!t]
% \caption{Overall results of \rag.}
% \label{tab:overall}
% \centering
% % \footnotesize
% \setlength{\tabcolsep}{3pt} % 调整列间距
% \fontsize{8}{9.6}\selectfont
% \begin{tabular}{lccccccc}
% \toprule
% & & \multicolumn{2}{c}{Single LLMs} & \multicolumn{2}{c}{General-Purpose Agent} & \multicolumn{2}{c}{\textbf{\rag (Ours)}} \\
% \cmidrule(lr){3-4} \cmidrule(lr){5-6} \cmidrule(lr){7-8}
% Category & Model & $R_{I}$ & $R_{E}$ & $R_{I}$ & $R_{E}$ & $R_{I}$ & $R_{E}$ \\
% \midrule
% \multirow{2}{*}{General LLMs} 
% & Qwen2.5-32B    & 17.52 & 9.52 & 8.11 & 5.47 & \textbf{19.68} & \textbf{12.78} \\
% & DeepSeek-V3    & 24.71 & 17.99 & 13.25 & 10.28 & \textbf{26.62} & \textbf{21.34} \\
% \midrule
% Domain LLMs 
% & DianJin-R1-32B & 21.94 & 17.84 & 11.17 & 10.69 & \textbf{23.87} & \textbf{18.87} \\
% \midrule
% \multirow{2}{*}{Reasoning-Augmented LLMs} 
% & QwQ-32B        & 23.63 & 18.80 & 12.42 & 11.03 & \textbf{26.46} & \textbf{22.15} \\
% & DeepSeek-R1    & 26.20 & 20.72 & 16.53 & 12.74 & \textbf{28.91} & \textbf{23.29} \\
% \bottomrule
% \end{tabular}
% \end{table}

To rigorously evaluate model performance, we define metrics at two levels of granularity: Recall at Issue level ($R_{I}$) and Recall at Evidence level ($R_{E}$). At the issue level, For each case $i$, let $S_i=\{s_{i}^{j}\}_{j=1}^{q_{i}}$ and $\hat{S}_i=\{\hat{s}_{i}^{j}\}_{j=1}^{\hat{q}_{i}}$ denote the sets of ground-truth and predicted fraud issues, respectively. We define a binary indicator function $\mathbb{I}_S(s, \hat{s})$ which takes value 1 if $s$ and $\hat{s}$ are judged semantically equivalent. The issue-level recall is then computed as: $R_{I} = \frac{1}{N} \sum_{i=1}^{N} \frac{\sum_{s \in S_{i}} \max_{\hat{s} \in \hat{S_{i}}} \mathbb{I}_{S}(s, \hat{s})}{|S_{i}|},$ which measures the average proportion of true fraud issues successfully matched by at least one predicted issue across all cases. At a finer granularity, for each matched issue pair $(s, \hat{s^{*}})$, we further assess the alignment of supporting evidence. Let $E_{s}$ and $\hat{E}_{\hat{s^{*}}}$ be the sets of ground-truth and predicted evidence points associated with issue $s$ and $\hat{s^{*}}$, respectively, and we define $\mathbb{I}_E(e, \hat{e})$ analogously to indicate semantic equivalence at the evidence level. The evidence-level recall is then formulated as: $R_{E} = \frac{1}{N} \sum_{i=1}^{N} \frac{1}{|S_{i}|} \sum_{s \in S_{i}} \left( \frac{\sum_{e \in E_s} \max_{\hat{e} \in \hat{E}_{\hat{s^{*}}}} \mathbb{I}_E(e, \hat{e})}{|E_s|} \right),$ which quantifies the extent to which the system retrieves both relevant issues and evidence for correctly identified fraud issues.

\subsection{Results}
\subsubsection{Overall Performance}
% \subsection{Effectiveness of \rag}
%为了对比体现我们方法的有效性，我们选择了两大类baseline：（1）单一长文本模型（通过截断上下文保证推理的完整性） （2）基于general-purpose agent框架例如React，让模型自主决定按照什么策略搜索造假证据链。在基座模型的选择上，我们选择了不同尺寸的推理模型与非推理模型覆盖了Deepseek R1以及QwQ 32B等前沿模型。 总的结果来看，general-purpose agent相较于single model的截断策略并没有显著提升风险精准召回率，可能的原因是因为大模型用于规划此类任务通常会随机生成搜索路径，与风险的实际先验分布不匹配导致大量无效信息干扰了模型的推理。而我们的方法，通过提前建模风险先验，并基于先验约束大模型的搜索路径从而实现召回率的大幅提升。除了召回精度，通过对模型搜索进行先验约束，我们的方法在推理效率以及token消耗上也大幅度优于普通的智能体范式。

% Overall, our results show that General-Purpose Agent paradigms do not yield significant improvements in precise risk recall compared to the Single LLMs with truncation strategy. A possible explanation is that LLMs, when tasked with planning in such scenarios, often generate limited common search paths, which tend to misalign with the actual prior distribution of fraud risks (shown in Fig \ref{fig:case}(b)). This mismatch introduces considerable irrelevant information, hindering effective reasoning. In contrast,  our method explicitly models risk priors in advance and constrains the model's search trajectory according to these priors, resulting in a substantial boost in both issue and evidence recall. %添加数值描述
\nosection{Basic comparison} 
Quantitative results in Table~\ref{tab:overall} demonstrate \rag's superior performance across both baseline categories. Compared to Single LLMs baseline, our framework achieves consistent improvements: For Reasoning-Augmented LLMs, \rag shows \textbf{10.3\%} higher issue-level recall ($R_I$) and \textbf{12.4\%} higher evidence-level recall ($R_E$) over DeepSeek-R1. The advantage becomes more pronounced against General-Purpose Agent baseline, with \textbf{74.8\%} and \textbf{82.8\%} relative improvements in $R_I$ and $R_E$ respectively. For General LLMs like DeepSeek-V3, \rag maintains \textbf{7.7\%} ($R_I$) and \textbf{18.6\%} ($R_E$) improvements over Single LLMs baseline, while achieving \textbf{100.9\%} and \textbf{107.6\%} greater performance than General-Purpose Agents on these metrics. These substantial margins highlight our architecture's effectiveness in coordinating specialized agents compared to both truncated context processing and undirected retrieval strategies.

Notably, our results show that General-Purpose Agent paradigms do not yield significant improvements in precise risk recall compared to the Single LLMs with truncation strategy. A possible explanation is that LLMs, when tasked with planning in such scenarios, often generate limited common search paths, which tend to misalign with the actual prior distribution of fraud risks (shown in Fig \ref{fig:case}(b)). This mismatch introduces considerable irrelevant information, hindering effective reasoning. In contrast,  our method explicitly models risk priors in advance and constrains the model's search trajectory according to these priors, resulting in a substantial boost in both issue and evidence recall. 

% The advantages of our framework extend consistently across distinct model types. For Reasoning-Augmented LLMs like QwQ-32B, \rag demonstrates improvements in both granularity levels through effective orchestration of multi-step reasoning processes. When applied to General LLMs such as Qwen2.5-32B, our risk-prioritized retrieval and collaborative verification mechanisms enhance the discovery of concealed fraud patterns that standard single-model approaches often overlook. This cross-category superiority underscores the generalizability of our multi-agent paradigm in synthesizing domain knowledge and distributed evidence.

\begin{figure} % 去掉星号 *
    \centering
    \includegraphics[width=\columnwidth]{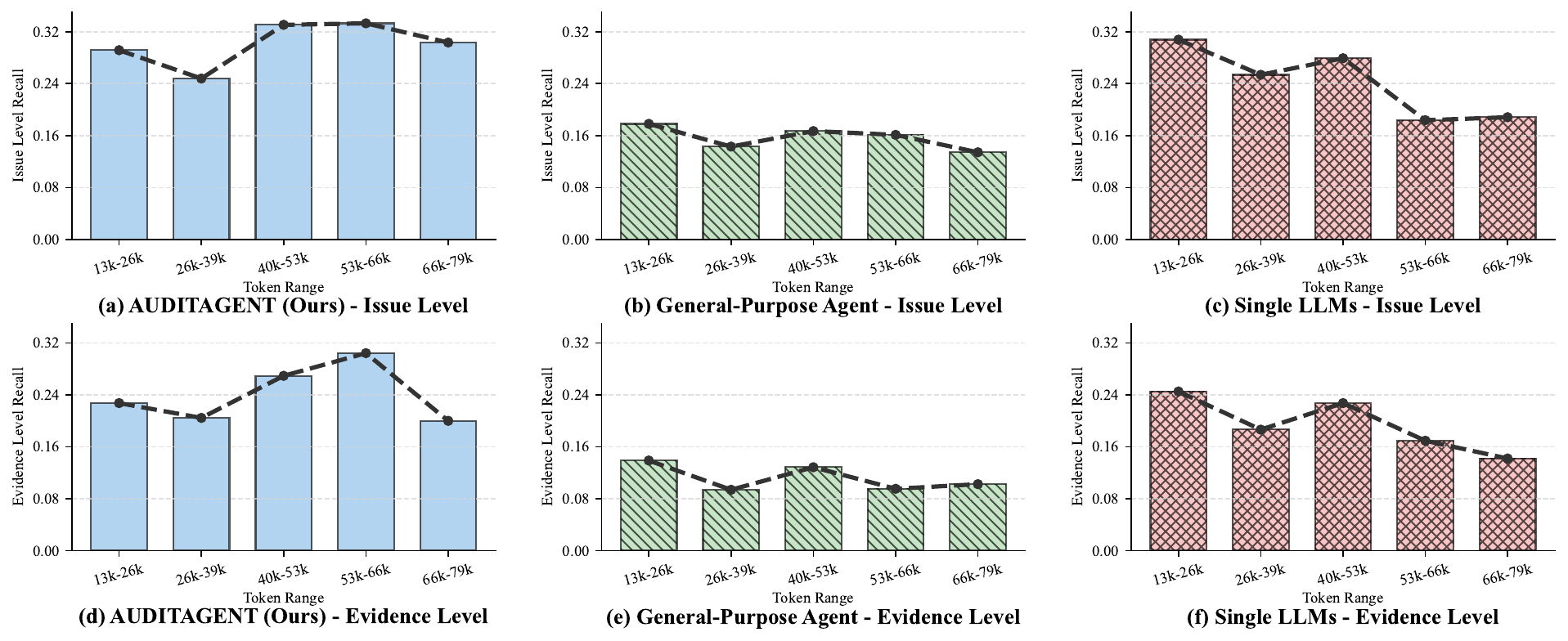} % 使用 \columnwidth
    \caption{Performance breakdown over cases with different number of input tokens.}
    \label{fig:tokens}
\end{figure}

\nosection{Performance comparison on various source lengths}
We divide the test set into five groups according to the number of input tokens in the financial reports and presents the recall achieved by each method on these subsets. As shown in Figure \ref{fig:tokens}, agent-based approaches that autonomously plan their reasoning process (\rag and General-Purpose Agent) achieve relatively stable performance across different groups. Notably, our method, which incorporates prior knowledge and expert insights, consistently outperforms all baseline methods by a significant margin. In contrast, Single LLMs approaches that rely on input truncation exhibit a clear decline in recall as the input token increases. These results further demonstrate the advantage of workflow designs that leverage prior knowledge for this challenging task.

% \begin{table*}
% \centering
% \caption{Performance comparison of Single-LLMs with different components.}
% \setlength{\tabcolsep}{3pt} % 调整列间距
% % \fontsize{8}{9.6}\selectfont
% \label{tab:comparison}
% \begin{tabular}{lccccccc}
% \toprule
%  & & \multicolumn{2}{c}{SL} & \multicolumn{2}{c}{SL w/ prior} & \multicolumn{2}{c}{SL w/ prior \& MoE} \\
% \cmidrule(lr){3-4} \cmidrule(lr){5-6} \cmidrule(lr){7-8}
% Category & Model & $R_I$ & $R_E$ & $R_I$ & $R_E$ & $R_I$ & $R_E$ \\
% \midrule
% \multirow{2}{*}{General LLMs} & Qwen2.5-32B & 17.52 & 9.52 & 18.36 & 10.23 & \textbf{19.68} & \textbf{12.78} \\
%  & DeepSeek-V3 & 24.71 & 17.99 & 25.43 & 18.86 & \textbf{26.62} & \textbf{21.34} \\
% \midrule
% Domain LLMs & Dianjin-R1-32B & 21.94 & 17.84 & \textbf{24.26} & \textbf{19.02} & 23.87 & 18.87 \\
% \midrule
% \multirow{2}{*}{Reasoning-Augmented LLMs} & QwQ-32B & 23.63 & 18.80 & 24.51 & 20.15 & \textbf{26.46} & \textbf{22.15} \\
%  & DeepSeek-R1 & 26.20 & 20.72 & 27.62 & 22.01 & \textbf{28.91} & \textbf{23.29} \\
% \bottomrule
% \end{tabular}
% \end{table*}

\begin{figure*}
    \centering
    \includegraphics[width=\textwidth]{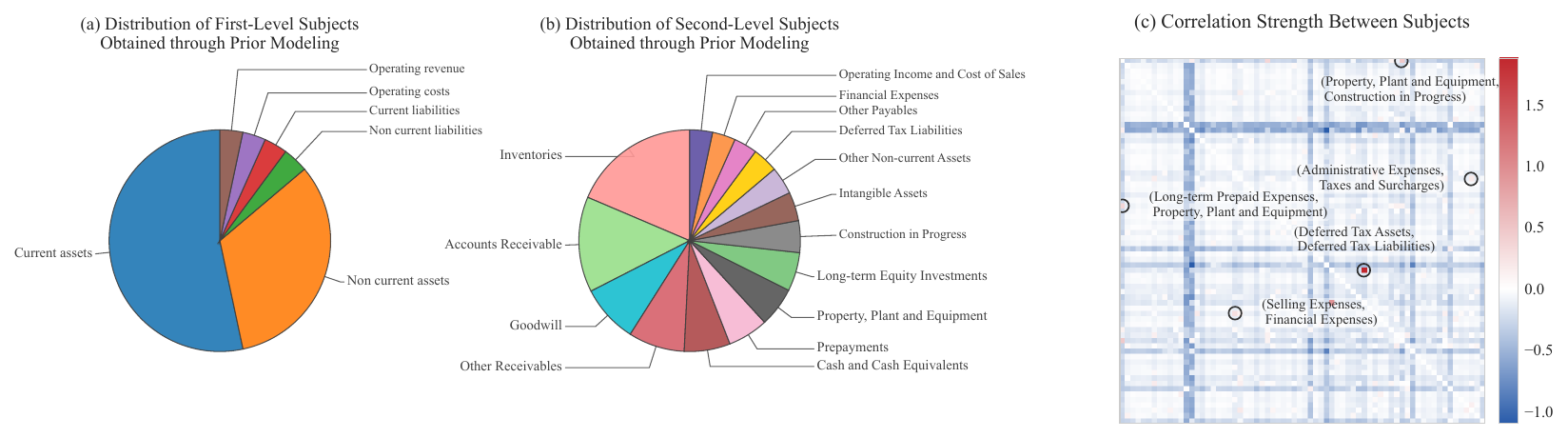}
    \caption{Fraudulent prior distribution over historical A-share data obtained by Subject Prior Modeling.}
    \label{fig:prior}
\end{figure*}

\subsubsection{Effectiveness of Subject Prior Modeling}
To further elucidate the effectiveness of our approach, we conduct a visual analysis of the fraud behavior priors extracted in the preliminary stage. Specifically, by modeling historical fraud cases with a probabilistic graphical model, we visualize in Figure \ref{fig:prior} the likelihood of fraudulent activities associated with both primary and secondary accounting categories. Moreover, leveraging the strengths of the probabilistic graphical framework, we are also able to identify pairs of accounting subjects that exhibit strong potential correlations in fraudulent schemes.
As shown in Figure \ref{fig:prior} (a), current and non-current assets emerge as the most important targets of fraudulent activities among primary accounting categories. In Figure \ref{fig:prior} (b), secondary categories such as goodwill and accounts receivable—items that are more susceptible to manipulation—are revealed as preferred targets for fraudsters. Furthermore, Figure \ref{fig:prior} (c) illustrates the distribution of the most importantly co-occurring subject pairs in historical fraud cases. These prior insights enable our framework to capture subtle risk patterns, particularly in detecting sophisticated fraud schemes like the association between construction-in-progress and fixed assets.

Building upon these fraud-related priors, we identified \textbf{15} critical accounting subjects that warrant focused attention. In contrast, frameworks without prior modeling typically consider all about \textbf{60-80} accounting subjects present in financial statements as potential risk indicators. Experimental results demonstrate the effectiveness of our approach: when implemented on the DeepSeek-R1 base model and Single LLMs framework, the prior modeling module achieves significant improvements of \textbf{5.4\%} and \textbf{6.2\%} in $R_{I}$ and $R_{E}$ metrics respectively, despite utilizing fewer contextual information of subjects, as detailed in Appendix. This performance gain highlights the importance of prior modeling, as models lacking this component tend to lose focus on critical risk indicators amidst excessive candidate subjects, ultimately compromising their ability to detect systematic financial fraud.
% Baseline experiments demonstrate that without prior-guided retrieval mechanisms, financial fraud cases typically involve 60-80 accounting subjects ($N_{\text{subjects}} \in [60,80]$), causing models to lose focus on critical risk subjects amidst excessive candidates and consequently fail to detect systematic financial fraud. The proposed Prior-Guided Multi-Route Retrieval framework achieves \textbf{5.4\%}  and \textbf{6.2\%} significant improvements in evidence chain integrity ($R_I$) and fraud detection accuracy ($R_E$) respectively when implemented on the Deepseek-R1 base model, as detailed in Appendix A.3. This enhancement originates from the prior-driven hierarchical retrieval strategy that enables compression of candidate subject space to high-probability risk subsets ($|\mathcal{S}_{\text{high-risk}}| \leq 15$) and establishment of cross-subject Bayesian inference pathways.

% \subsubsection{Effectiveness of Multi-Expert Reasoning}
%Beyond recall performance, our approach also demonstrates notable advantages in inference efficiency and token consumption. By imposing prior-informed constraints on model search, we achieve more efficient and targeted reasoning than conventional agent-based paradigms.%添加数值描述

\subsubsection{Impact of Different Base Models}
\begin{figure*}
    \centering
    \includegraphics[width=0.9\textwidth]{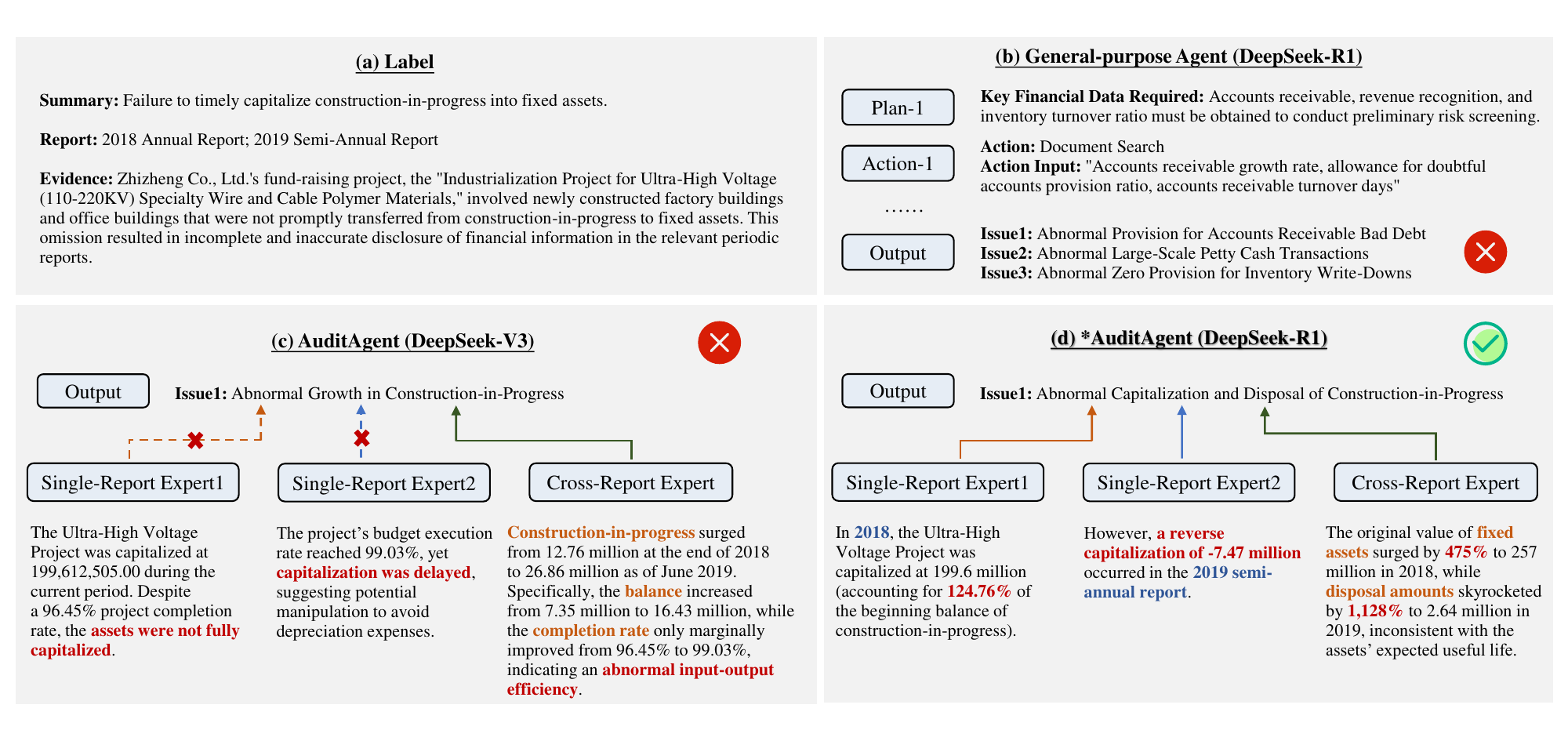}
    \caption{Case Study to show (1) The superiority of \rag comparing with General-Purpose Agent; (2) The superiority of r1-like reasoning models for analyzing evidences.}
    \label{fig:case}
\end{figure*}

\nosection{Can slow-thinking capabilities improve \rag's performance}
Reasoning models equipped with slow-thinking capabilities, such as DeepSeek-R1, demonstrate a significant advantage over conventional LLMs. For instance, within our \rag framework, DeepSeek-R1 improves \( R_I \) and \( R_E \) from 26.62\% and 21.34\% (as achieved by the General LLM DeepSeek-V3) to 28.9\% and 23.29\%, respectively. Subsequent case studies (shown in Fig \ref{fig:case}(c) and Fig \ref{fig:case}(d)) reveal that reasoning models substantially outperform general models in integrating and associating subtle, contextually relevant clues.

\nosection{Can domain-specific LLMs optimized for the financial sector enhance \rag’s performance}
In addition, we also adapt the latest LLMs that are specifically optimized for the financial domain into our framework. As shown in Table \ref{tab:overall}, applying the latest financial domain-specific model, DianJin-R1-32B\cite{zhu2025dianjin}, to fraud clue analysis within our framework yields an 4.19\% and 6.09\% relative improvement in $R_{I}$ and $R_{E}$ compared to Qwen2.5-32B with the same parameter size. 

% Detailed case analysis indicates that this advantage primarily stems from the domain-specific model’s superior capability in first-order financial data reasoning tasks.
% \section{Result}

%\nosection{Effectiveness in handling multi-reports}

%对比单份财报，多份财报结果

% \subsection{Insights from Fraudulent Prior}
%为了进一步理解我们的方法的有效性，我们对前置步骤提取的造假行为先验进行了可视化分析。具体来说，基于对历史造假案例进行概率图模型建模，我们在图x中展示了一级科目与二级子科目的造假可能性，除此之外，得益于概率图模型的优势，我们也可以获取关联性较高的潜在造假科目对。从结果可以看出，

% \section{Result}

\section{Conclusion}
In this work, we propose a novel framework for financial fraud detection that integrates domain-specific risk priors and expert knowledge into a multi-agent large language model reasoning architecture. Through comprehensive experiments on a real-world dataset of financial reports and enforcement cases, we demonstrate that our approach significantly improves both the precision and recall of fraud risk identification compared to General-Purpose Agent framework and Single-LLMs baselines. By explicitly modeling the prior distribution of fraudulent behaviors and constraining the reasoning process accordingly, our method achieves more accurate evidence localization, enhanced interpretability, and greater inference efficiency. Further case studies highlight the superior capability of reasoning-augmented models in synthesizing and connecting subtle clues across complex financial documents. Our findings suggest that the incorporation of domain knowledge and structured priors is vital for advancing automated, fine-grained financial risk analysis. Future work will explore the application of our paradigm to broader risk domains and the integration of interactive expert-in-the-loop learning for continual improvement.

% \section{Acknowledgments}

% Identification of funding sources and other support, and thanks to
% individuals and groups that assisted in the research and the
% preparation of the work should be included in an acknowledgment
% section, which is placed just before the reference section in your
% document.

%%
%% The next two lines define the bibliography style to be used, and
%% the bibliography file.
\bibliographystyle{ACM-Reference-Format}
% \bibliography{sample-base}

% \bibliographystyle{unsrtnat}
\bibliography{reference.bib} % 指向bib文件

%%% -*-BibTeX-*-
%%% Do NOT edit. File created by BibTeX with style
%%% ACM-Reference-Format-Journals [18-Jan-2012].

\begin{thebibliography}{50}

%%% ====================================================================
%%% NOTE TO THE USER: you can override these defaults by providing
%%% customized versions of any of these macros before the \bibliography
%%% command.  Each of them MUST provide its own final punctuation,
%%% except for \shownote{} and \showURL{}.  The latter two
%%% do not use final punctuation, in order to avoid confusing it with
%%% the Web address.
%%%
%%% To suppress output of a particular field, define its macro to expand
%%% to an empty string, or better, \unskip, like this:
%%%
%%% \newcommand{\showURL}[1]{\unskip}   % LaTeX syntax
%%%
%%% \def \showURL #1{\unskip}           % plain TeX syntax
%%%
%%% ====================================================================

\ifx \showCODEN    \undefined \def \showCODEN     #1{\unskip}     \fi
\ifx \showISBNx    \undefined \def \showISBNx     #1{\unskip}     \fi
\ifx \showISBNxiii \undefined \def \showISBNxiii  #1{\unskip}     \fi
\ifx \showISSN     \undefined \def \showISSN      #1{\unskip}     \fi
\ifx \showLCCN     \undefined \def \showLCCN      #1{\unskip}     \fi
\ifx \shownote     \undefined \def \shownote      #1{#1}          \fi
\ifx \showarticletitle \undefined \def \showarticletitle #1{#1}   \fi
\ifx \showURL      \undefined \def \showURL       {\relax}        \fi
% The following commands are used for tagged output and should be
% invisible to TeX
\providecommand\bibfield[2]{#2}
\providecommand\bibinfo[2]{#2}
\providecommand\natexlab[1]{#1}
\providecommand\showeprint[2][]{arXiv:#2}

\bibitem[csr(2008)]%
        {csrcCSRC}
 \bibinfo{year}{2008}\natexlab{}.
\newblock \bibinfo{title}{{C}{S}{R}{C} --- csrc.gov.cn}.
\newblock \bibinfo{howpublished}{\url{http://www.csrc.gov.cn/csrc_en/index.shtml}}.
\newblock


\bibitem[git(2025)]%
        {githubGitHubPymupdfPyMuPDF}
 \bibinfo{year}{2025}\natexlab{}.
\newblock \bibinfo{title}{{G}it{H}ub - pymupdf/{P}y{M}u{P}{D}{F}: {P}y{M}u{P}{D}{F} is a high performance {P}ython library for data extraction, analysis, conversion \& manipulation of {P}{D}{F} (and other) documents. --- github.com}.
\newblock \bibinfo{howpublished}{\url{https://github.com/pymupdf/PyMuPDF}}.
\newblock


\bibitem[Ali et~al\mbox{.}(2022)]%
        {ali2022financial}
\bibfield{author}{\bibinfo{person}{Abdulalem Ali}, \bibinfo{person}{Shukor Abd~Razak}, \bibinfo{person}{Siti~Hajar Othman}, \bibinfo{person}{Taiseer Abdalla~Elfadil Eisa}, \bibinfo{person}{Arafat Al-Dhaqm}, \bibinfo{person}{Maged Nasser}, \bibinfo{person}{Tusneem Elhassan}, \bibinfo{person}{Hashim Elshafie}, {and} \bibinfo{person}{Abdu Saif}.} \bibinfo{year}{2022}\natexlab{}.
\newblock \showarticletitle{Financial fraud detection based on machine learning: a systematic literature review}.
\newblock \bibinfo{journal}{\emph{Applied Sciences}} \bibinfo{volume}{12}, \bibinfo{number}{19} (\bibinfo{year}{2022}), \bibinfo{pages}{9637}.
\newblock


\bibitem[Apriliana and Agustina(2017)]%
        {apriliana2017analysis}
\bibfield{author}{\bibinfo{person}{Siska Apriliana} {and} \bibinfo{person}{Linda Agustina}.} \bibinfo{year}{2017}\natexlab{}.
\newblock \showarticletitle{The analysis of fraudulent financial reporting determinant through fraud pentagon approach}.
\newblock \bibinfo{journal}{\emph{Jurnal Dinamika Akuntansi}} \bibinfo{volume}{9}, \bibinfo{number}{2} (\bibinfo{year}{2017}), \bibinfo{pages}{154--165}.
\newblock


\bibitem[Ashtiani and Raahemi(2021)]%
        {ashtiani2021intelligent}
\bibfield{author}{\bibinfo{person}{Matin~N Ashtiani} {and} \bibinfo{person}{Bijan Raahemi}.} \bibinfo{year}{2021}\natexlab{}.
\newblock \showarticletitle{Intelligent fraud detection in financial statements using machine learning and data mining: a systematic literature review}.
\newblock \bibinfo{journal}{\emph{Ieee Access}}  \bibinfo{volume}{10} (\bibinfo{year}{2021}), \bibinfo{pages}{72504--72525}.
\newblock


\bibitem[Bhattacharya and Mickovic(2024)]%
        {IJAIS_binary_FinFraud}
\bibfield{author}{\bibinfo{person}{Indranil Bhattacharya} {and} \bibinfo{person}{Ana Mickovic}.} \bibinfo{year}{2024}\natexlab{}.
\newblock \showarticletitle{Accounting fraud detection using contextual language learning}.
\newblock \bibinfo{journal}{\emph{International Journal of Accounting Information Systems}}  \bibinfo{volume}{53} (\bibinfo{year}{2024}), \bibinfo{pages}{100682}.
\newblock


\bibitem[Brazel et~al\mbox{.}(2015)]%
        {brazel2015understanding}
\bibfield{author}{\bibinfo{person}{Joseph~F Brazel}, \bibinfo{person}{Keith~L Jones}, \bibinfo{person}{Jane Thayer}, {and} \bibinfo{person}{Rick~C Warne}.} \bibinfo{year}{2015}\natexlab{}.
\newblock \showarticletitle{Understanding investor perceptions of financial statement fraud and their use of red flags: Evidence from the field}.
\newblock \bibinfo{journal}{\emph{Review of Accounting Studies}}  \bibinfo{volume}{20} (\bibinfo{year}{2015}), \bibinfo{pages}{1373--1406}.
\newblock


\bibitem[Campa et~al\mbox{.}(2023)]%
        {campa2023roles}
\bibfield{author}{\bibinfo{person}{Domenico Campa}, \bibinfo{person}{Alberto Quagli}, {and} \bibinfo{person}{Paola Ramassa}.} \bibinfo{year}{2023}\natexlab{}.
\newblock \showarticletitle{The roles and interplay of enforcers and auditors in the context of accounting fraud: a review of the accounting literature}.
\newblock \bibinfo{journal}{\emph{Journal of Accounting Literature}} \bibinfo{volume}{47}, \bibinfo{number}{5} (\bibinfo{year}{2023}), \bibinfo{pages}{151--183}.
\newblock


\bibitem[Chen et~al\mbox{.}(2025)]%
        {chen2025towards}
\bibfield{author}{\bibinfo{person}{Qiguang Chen}, \bibinfo{person}{Libo Qin}, \bibinfo{person}{Jinhao Liu}, \bibinfo{person}{Dengyun Peng}, \bibinfo{person}{Jiannan Guan}, \bibinfo{person}{Peng Wang}, \bibinfo{person}{Mengkang Hu}, \bibinfo{person}{Yuhang Zhou}, \bibinfo{person}{Te Gao}, {and} \bibinfo{person}{Wangxiang Che}.} \bibinfo{year}{2025}\natexlab{}.
\newblock \showarticletitle{Towards reasoning era: A survey of long chain-of-thought for reasoning large language models}.
\newblock \bibinfo{journal}{\emph{arXiv preprint arXiv:2503.09567}} (\bibinfo{year}{2025}).
\newblock


\bibitem[Chen et~al\mbox{.}(2021)]%
        {chen2021finqa}
\bibfield{author}{\bibinfo{person}{Zhiyu Chen}, \bibinfo{person}{Wenhu Chen}, \bibinfo{person}{Charese Smiley}, \bibinfo{person}{Sameena Shah}, \bibinfo{person}{Iana Borova}, \bibinfo{person}{Dylan Langdon}, \bibinfo{person}{Reema Moussa}, \bibinfo{person}{Matt Beane}, \bibinfo{person}{Ting-Hao Huang}, \bibinfo{person}{Bryan Routledge}, {and} \bibinfo{person}{William~Yang Wang}.} \bibinfo{year}{2021}\natexlab{}.
\newblock \showarticletitle{{F}in{QA}: A Dataset of Numerical Reasoning over Financial Data}. In \bibinfo{booktitle}{\emph{Proceedings of the 2021 Conference on Empirical Methods in Natural Language Processing}}. \bibinfo{publisher}{Association for Computational Linguistics}.
\newblock


\bibitem[Chen et~al\mbox{.}(2022)]%
        {chen2022convfinqa}
\bibfield{author}{\bibinfo{person}{Zhiyu Chen}, \bibinfo{person}{Shiyang Li}, \bibinfo{person}{Charese Smiley}, \bibinfo{person}{Zhiqiang Ma}, \bibinfo{person}{Sameena Shah}, {and} \bibinfo{person}{William~Yang Wang}.} \bibinfo{year}{2022}\natexlab{}.
\newblock \showarticletitle{{C}onv{F}in{QA}: Exploring the Chain of Numerical Reasoning in Conversational Finance Question Answering}. In \bibinfo{booktitle}{\emph{Proceedings of the 2022 Conference on Empirical Methods in Natural Language Processing}}. \bibinfo{publisher}{Association for Computational Linguistics}.
\newblock


\bibitem[Craja et~al\mbox{.}(2020)]%
        {DSS_metric_FinFraud}
\bibfield{author}{\bibinfo{person}{Patricia Craja}, \bibinfo{person}{Alisa Kim}, {and} \bibinfo{person}{Stefan Lessmann}.} \bibinfo{year}{2020}\natexlab{}.
\newblock \showarticletitle{Deep learning for detecting financial statement fraud}.
\newblock \bibinfo{journal}{\emph{Decision Support Systems}}  \bibinfo{volume}{139} (\bibinfo{year}{2020}), \bibinfo{pages}{113421}.
\newblock


\bibitem[Elsayed(2017)]%
        {elsayed2017indicators}
\bibfield{author}{\bibinfo{person}{Ashraf Elsayed}.} \bibinfo{year}{2017}\natexlab{}.
\newblock \showarticletitle{Indicators of the financial statement fraud (red flags)}.
\newblock \bibinfo{journal}{\emph{Available at SSRN 3074187}} (\bibinfo{year}{2017}).
\newblock


\bibitem[Glancy and Yadav(2011)]%
        {DSS_Text_FinFraud}
\bibfield{author}{\bibinfo{person}{Fletcher~H. Glancy} {and} \bibinfo{person}{Surya~B. Yadav}.} \bibinfo{year}{2011}\natexlab{}.
\newblock \showarticletitle{A computational model for financial reporting fraud detection}.
\newblock \bibinfo{journal}{\emph{Decision Support Systems}} \bibinfo{volume}{50}, \bibinfo{number}{3} (\bibinfo{year}{2011}), \bibinfo{pages}{595--601}.
\newblock


\bibitem[Guo et~al\mbox{.}(2025b)]%
        {guo2025deepseek}
\bibfield{author}{\bibinfo{person}{Daya Guo}, \bibinfo{person}{Dejian Yang}, \bibinfo{person}{Haowei Zhang}, \bibinfo{person}{Junxiao Song}, \bibinfo{person}{Ruoyu Zhang}, \bibinfo{person}{Runxin Xu}, \bibinfo{person}{Qihao Zhu}, \bibinfo{person}{Shirong Ma}, \bibinfo{person}{Peiyi Wang}, \bibinfo{person}{Xiao Bi}, {et~al\mbox{.}}} \bibinfo{year}{2025}\natexlab{b}.
\newblock \showarticletitle{Deepseek-r1: Incentivizing reasoning capability in llms via reinforcement learning}.
\newblock \bibinfo{journal}{\emph{arXiv preprint arXiv:2501.12948}} (\bibinfo{year}{2025}).
\newblock


\bibitem[Guo et~al\mbox{.}(2025a)]%
        {zhang2023fineval}
\bibfield{author}{\bibinfo{person}{Xin Guo}, \bibinfo{person}{Haotian Xia}, \bibinfo{person}{Zhaowei Liu}, \bibinfo{person}{Hanyang Cao}, \bibinfo{person}{Zhi Yang}, \bibinfo{person}{Zhiqiang Liu}, \bibinfo{person}{Sizhe Wang}, \bibinfo{person}{Jinyi Niu}, \bibinfo{person}{Chuqi Wang}, \bibinfo{person}{Yanhui Wang}, \bibinfo{person}{Xiaolong Liang}, \bibinfo{person}{Xiaoming Huang}, \bibinfo{person}{Bing Zhu}, \bibinfo{person}{Zhongyu Wei}, \bibinfo{person}{Yun Chen}, \bibinfo{person}{Weining Shen}, {and} \bibinfo{person}{Liwen Zhang}.} \bibinfo{year}{2025}\natexlab{a}.
\newblock \showarticletitle{{F}in{E}val: A {C}hinese Financial Domain Knowledge Evaluation Benchmark for Large Language Models}. In \bibinfo{booktitle}{\emph{Proceedings of the 2025 Conference of the Nations of the Americas Chapter of the Association for Computational Linguistics: Human Language Technologies (Volume 1: Long Papers)}}. \bibinfo{publisher}{Association for Computational Linguistics}.
\newblock


\bibitem[Hilal et~al\mbox{.}(2022)]%
        {hilal2022financial}
\bibfield{author}{\bibinfo{person}{Waleed Hilal}, \bibinfo{person}{S~Andrew Gadsden}, {and} \bibinfo{person}{John Yawney}.} \bibinfo{year}{2022}\natexlab{}.
\newblock \showarticletitle{Financial fraud: a review of anomaly detection techniques and recent advances}.
\newblock \bibinfo{journal}{\emph{Expert systems With applications}}  \bibinfo{volume}{193} (\bibinfo{year}{2022}), \bibinfo{pages}{116429}.
\newblock


\bibitem[Karanikolas et~al\mbox{.}(2023)]%
        {karanikolas2023large}
\bibfield{author}{\bibinfo{person}{Nikitas Karanikolas}, \bibinfo{person}{Eirini Manga}, \bibinfo{person}{Nikoletta Samaridi}, \bibinfo{person}{Eleni Tousidou}, {and} \bibinfo{person}{Michael Vassilakopoulos}.} \bibinfo{year}{2023}\natexlab{}.
\newblock \showarticletitle{Large language models versus natural language understanding and generation}. In \bibinfo{booktitle}{\emph{Proceedings of the 27th Pan-Hellenic Conference on Progress in Computing and Informatics}}. \bibinfo{pages}{278--290}.
\newblock


\bibitem[Krumdick et~al\mbox{.}(2024)]%
        {koncel2023bizbench}
\bibfield{author}{\bibinfo{person}{Michael Krumdick}, \bibinfo{person}{Rik Koncel-Kedziorski}, \bibinfo{person}{Viet~Dac Lai}, \bibinfo{person}{Varshini Reddy}, \bibinfo{person}{Charles Lovering}, {and} \bibinfo{person}{Chris Tanner}.} \bibinfo{year}{2024}\natexlab{}.
\newblock \showarticletitle{{B}iz{B}ench: A Quantitative Reasoning Benchmark for Business and Finance}. In \bibinfo{booktitle}{\emph{Proceedings of the 62nd Annual Meeting of the Association for Computational Linguistics (Volume 1: Long Papers)}}. \bibinfo{publisher}{Association for Computational Linguistics}, \bibinfo{address}{Bangkok, Thailand}.
\newblock


\bibitem[Kuratov et~al\mbox{.}(2024)]%
        {NEURIPS2024_c0d62e70}
\bibfield{author}{\bibinfo{person}{Yuri Kuratov}, \bibinfo{person}{Aydar Bulatov}, \bibinfo{person}{Petr Anokhin}, \bibinfo{person}{Ivan Rodkin}, \bibinfo{person}{Dmitry Sorokin}, \bibinfo{person}{Artyom Sorokin}, {and} \bibinfo{person}{Mikhail Burtsev}.} \bibinfo{year}{2024}\natexlab{}.
\newblock \showarticletitle{BABILong: Testing the Limits of LLMs with Long Context Reasoning-in-a-Haystack}. In \bibinfo{booktitle}{\emph{Advances in Neural Information Processing Systems}}, Vol.~\bibinfo{volume}{37}. \bibinfo{pages}{106519--106554}.
\newblock


\bibitem[Li et~al\mbox{.}(2025)]%
        {li2025search}
\bibfield{author}{\bibinfo{person}{Xiaoxi Li}, \bibinfo{person}{Guanting Dong}, \bibinfo{person}{Jiajie Jin}, \bibinfo{person}{Yuyao Zhang}, \bibinfo{person}{Yujia Zhou}, \bibinfo{person}{Yutao Zhu}, \bibinfo{person}{Peitian Zhang}, {and} \bibinfo{person}{Zhicheng Dou}.} \bibinfo{year}{2025}\natexlab{}.
\newblock \showarticletitle{Search-o1: Agentic search-enhanced large reasoning models}.
\newblock \bibinfo{journal}{\emph{arXiv preprint arXiv:2501.05366}} (\bibinfo{year}{2025}).
\newblock


\bibitem[Lin and Gao(2022)]%
        {ESWA_interpretability_FinFraud}
\bibfield{author}{\bibinfo{person}{Kang Lin} {and} \bibinfo{person}{Yuzhuo Gao}.} \bibinfo{year}{2022}\natexlab{}.
\newblock \showarticletitle{Model interpretability of financial fraud detection by group SHAP}.
\newblock \bibinfo{journal}{\emph{Expert Systems with Applications}}  \bibinfo{volume}{210} (\bibinfo{year}{2022}), \bibinfo{pages}{118354}.
\newblock


\bibitem[Liu et~al\mbox{.}(2024a)]%
        {liu2024deepseek}
\bibfield{author}{\bibinfo{person}{Aixin Liu}, \bibinfo{person}{Bei Feng}, \bibinfo{person}{Bing Xue}, \bibinfo{person}{Bingxuan Wang}, \bibinfo{person}{Bochao Wu}, \bibinfo{person}{Chengda Lu}, \bibinfo{person}{Chenggang Zhao}, \bibinfo{person}{Chengqi Deng}, \bibinfo{person}{Chenyu Zhang}, \bibinfo{person}{Chong Ruan}, {et~al\mbox{.}}} \bibinfo{year}{2024}\natexlab{a}.
\newblock \showarticletitle{Deepseek-v3 technical report}.
\newblock \bibinfo{journal}{\emph{arXiv preprint arXiv:2412.19437}} (\bibinfo{year}{2024}).
\newblock


\bibitem[Liu et~al\mbox{.}(2025)]%
        {liu2025comprehensive}
\bibfield{author}{\bibinfo{person}{Jiaheng Liu}, \bibinfo{person}{Dawei Zhu}, \bibinfo{person}{Zhiqi Bai}, \bibinfo{person}{Yancheng He}, \bibinfo{person}{Huanxuan Liao}, \bibinfo{person}{Haoran Que}, \bibinfo{person}{Zekun Wang}, \bibinfo{person}{Chenchen Zhang}, \bibinfo{person}{Ge Zhang}, \bibinfo{person}{Jiebin Zhang}, {et~al\mbox{.}}} \bibinfo{year}{2025}\natexlab{}.
\newblock \showarticletitle{A Comprehensive Survey on Long Context Language Modeling}.
\newblock \bibinfo{journal}{\emph{arXiv preprint arXiv:2503.17407}} (\bibinfo{year}{2025}).
\newblock


\bibitem[Liu et~al\mbox{.}(2024b)]%
        {liu-etal-2024-lost}
\bibfield{author}{\bibinfo{person}{Nelson~F. Liu}, \bibinfo{person}{Kevin Lin}, \bibinfo{person}{John Hewitt}, \bibinfo{person}{Ashwin Paranjape}, \bibinfo{person}{Michele Bevilacqua}, \bibinfo{person}{Fabio Petroni}, {and} \bibinfo{person}{Percy Liang}.} \bibinfo{year}{2024}\natexlab{b}.
\newblock \showarticletitle{Lost in the Middle: How Language Models Use Long Contexts}.
\newblock \bibinfo{journal}{\emph{Transactions of the Association for Computational Linguistics}}  \bibinfo{volume}{12} (\bibinfo{year}{2024}), \bibinfo{pages}{157--173}.
\newblock


\bibitem[Nesvijevskaia et~al\mbox{.}(2021)]%
        {nesvijevskaia2021accuracy}
\bibfield{author}{\bibinfo{person}{Anna Nesvijevskaia}, \bibinfo{person}{Sophie Ouillade}, \bibinfo{person}{Pauline Guilmin}, {and} \bibinfo{person}{Jean-Daniel Zucker}.} \bibinfo{year}{2021}\natexlab{}.
\newblock \showarticletitle{The accuracy versus interpretability trade-off in fraud detection model}.
\newblock \bibinfo{journal}{\emph{Data \& Policy}}  \bibinfo{volume}{3} (\bibinfo{year}{2021}), \bibinfo{pages}{e12}.
\newblock


\bibitem[Reurink(2019)]%
        {reurink2019financial}
\bibfield{author}{\bibinfo{person}{Arjan Reurink}.} \bibinfo{year}{2019}\natexlab{}.
\newblock \showarticletitle{Financial fraud: A literature review}.
\newblock \bibinfo{journal}{\emph{Contemporary topics in finance: A collection of literature surveys}} (\bibinfo{year}{2019}), \bibinfo{pages}{79--115}.
\newblock


\bibitem[S{\'a}nchez-Aguayo et~al\mbox{.}(2021)]%
        {sanchez2021fraud}
\bibfield{author}{\bibinfo{person}{Marco S{\'a}nchez-Aguayo}, \bibinfo{person}{Luis Urquiza-Aguiar}, {and} \bibinfo{person}{Jos{\'e} Estrada-Jim{\'e}nez}.} \bibinfo{year}{2021}\natexlab{}.
\newblock \showarticletitle{Fraud detection using the fraud triangle theory and data mining techniques: a literature review}.
\newblock \bibinfo{journal}{\emph{Computers}} \bibinfo{volume}{10}, \bibinfo{number}{10} (\bibinfo{year}{2021}), \bibinfo{pages}{121}.
\newblock


\bibitem[Shen et~al\mbox{.}(2023)]%
        {shen2023positional}
\bibfield{author}{\bibinfo{person}{Ruoqi Shen}, \bibinfo{person}{S{\'e}bastien Bubeck}, \bibinfo{person}{Ronen Eldan}, \bibinfo{person}{Yin~Tat Lee}, \bibinfo{person}{Yuanzhi Li}, {and} \bibinfo{person}{Yi Zhang}.} \bibinfo{year}{2023}\natexlab{}.
\newblock \showarticletitle{Positional description matters for transformers arithmetic}.
\newblock \bibinfo{journal}{\emph{arXiv preprint arXiv:2311.14737}} (\bibinfo{year}{2023}).
\newblock


\bibitem[(SSIC)(2025)]%
        {cninfox5DE8x6F6Ex8D44x8BAFx7F51}
\bibfield{author}{\bibinfo{person}{Shenzhen Securities Information Co.~Ltd. (SSIC)}.} \bibinfo{year}{2025}\natexlab{}.
\newblock \bibinfo{title}{cninfo.com.cn}.
\newblock \bibinfo{howpublished}{\url{https://www.cninfo.com.cn/new/index}}.
\newblock


\bibitem[Sun et~al\mbox{.}(2024)]%
        {sun-etal-2024-pearl}
\bibfield{author}{\bibinfo{person}{Simeng Sun}, \bibinfo{person}{Yang Liu}, \bibinfo{person}{Shuohang Wang}, \bibinfo{person}{Dan Iter}, \bibinfo{person}{Chenguang Zhu}, {and} \bibinfo{person}{Mohit Iyyer}.} \bibinfo{year}{2024}\natexlab{}.
\newblock \showarticletitle{{PEARL}: Prompting Large Language Models to Plan and Execute Actions Over Long Documents}. In \bibinfo{booktitle}{\emph{Proceedings of the 18th Conference of the European Chapter of the Association for Computational Linguistics (Volume 1: Long Papers)}}. \bibinfo{publisher}{Association for Computational Linguistics}, \bibinfo{pages}{469--486}.
\newblock


\bibitem[Team(2025)]%
        {qwenlmQwQReflect}
\bibfield{author}{\bibinfo{person}{Qwen Team}.} \bibinfo{year}{2025}\natexlab{}.
\newblock \bibinfo{title}{{Q}w{Q}: {R}eflect {D}eeply on the {B}oundaries of the {U}nknown}.
\newblock \bibinfo{howpublished}{\url{https://qwenlm.github.io/blog/qwq-32b-preview/}}.
\newblock


\bibitem[Umar et~al\mbox{.}(2020)]%
        {umar2020fraud}
\bibfield{author}{\bibinfo{person}{Haryono Umar}, \bibinfo{person}{Dantes Partahi}, {and} \bibinfo{person}{Rahima~Br Purba}.} \bibinfo{year}{2020}\natexlab{}.
\newblock \showarticletitle{Fraud diamond analysis in detecting fraudulent financial report}.
\newblock \bibinfo{journal}{\emph{International Journal of Scientific and Technology Research}} \bibinfo{volume}{9}, \bibinfo{number}{3} (\bibinfo{year}{2020}), \bibinfo{pages}{6638--6646}.
\newblock


\bibitem[Wang et~al\mbox{.}(2023)]%
        {DSS_multimodal_FinFraud}
\bibfield{author}{\bibinfo{person}{Gang Wang}, \bibinfo{person}{Jingling Ma}, {and} \bibinfo{person}{Gang Chen}.} \bibinfo{year}{2023}\natexlab{}.
\newblock \showarticletitle{Attentive statement fraud detection: Distinguishing multimodal financial data with fine-grained attention}.
\newblock \bibinfo{journal}{\emph{Decision Support Systems}}  \bibinfo{volume}{167} (\bibinfo{year}{2023}), \bibinfo{pages}{113913}.
\newblock


\bibitem[Wang et~al\mbox{.}(2024)]%
        {wang2024survey}
\bibfield{author}{\bibinfo{person}{Lei Wang}, \bibinfo{person}{Chen Ma}, \bibinfo{person}{Xueyang Feng}, \bibinfo{person}{Zeyu Zhang}, \bibinfo{person}{Hao Yang}, \bibinfo{person}{Jingsen Zhang}, \bibinfo{person}{Zhiyuan Chen}, \bibinfo{person}{Jiakai Tang}, \bibinfo{person}{Xu Chen}, \bibinfo{person}{Yankai Lin}, {et~al\mbox{.}}} \bibinfo{year}{2024}\natexlab{}.
\newblock \showarticletitle{A survey on large language model based autonomous agents}.
\newblock \bibinfo{journal}{\emph{Frontiers of Computer Science}} \bibinfo{volume}{18}, \bibinfo{number}{6} (\bibinfo{year}{2024}), \bibinfo{pages}{186345}.
\newblock


\bibitem[Xie et~al\mbox{.}(2024)]%
        {xie2024finben}
\bibfield{author}{\bibinfo{person}{Qianqian Xie}, \bibinfo{person}{Weiguang Han}, \bibinfo{person}{Zhengyu Chen}, \bibinfo{person}{Ruoyu Xiang}, \bibinfo{person}{Xiao Zhang}, \bibinfo{person}{Yueru He}, \bibinfo{person}{Mengxi Xiao}, \bibinfo{person}{Dong Li}, \bibinfo{person}{Yongfu Dai}, \bibinfo{person}{Duanyu Feng}, {et~al\mbox{.}}} \bibinfo{year}{2024}\natexlab{}.
\newblock \showarticletitle{Finben: A holistic financial benchmark for large language models}.
\newblock \bibinfo{journal}{\emph{Advances in Neural Information Processing Systems}}  \bibinfo{volume}{37} (\bibinfo{year}{2024}), \bibinfo{pages}{95716--95743}.
\newblock


\bibitem[Xin et~al\mbox{.}(2018)]%
        {xin2018economic}
\bibfield{author}{\bibinfo{person}{Qingquan Xin}, \bibinfo{person}{Jing Zhou}, {and} \bibinfo{person}{Fang Hu}.} \bibinfo{year}{2018}\natexlab{}.
\newblock \showarticletitle{The economic consequences of financial fraud: evidence from the product market in China}.
\newblock \bibinfo{journal}{\emph{China Journal of Accounting Studies}} \bibinfo{volume}{6}, \bibinfo{number}{1} (\bibinfo{year}{2018}), \bibinfo{pages}{1--23}.
\newblock


\bibitem[Xu et~al\mbox{.}(2022)]%
        {WWW_Elbo}
\bibfield{author}{\bibinfo{person}{Xiaoxiao Xu}, \bibinfo{person}{Chen Yang}, \bibinfo{person}{Qian Yu}, \bibinfo{person}{Zhiwei Fang}, \bibinfo{person}{Jiaxing Wang}, \bibinfo{person}{Chaosheng Fan}, \bibinfo{person}{Yang He}, \bibinfo{person}{Changping Peng}, \bibinfo{person}{Zhangang Lin}, {and} \bibinfo{person}{Jingping Shao}.} \bibinfo{year}{2022}\natexlab{}.
\newblock \showarticletitle{Alleviating Cold-start Problem in CTR Prediction with A Variational Embedding Learning Framework}. In \bibinfo{booktitle}{\emph{Proceedings of the ACM Web Conference 2022}}. \bibinfo{publisher}{Association for Computing Machinery}, \bibinfo{pages}{27–35}.
\newblock


\bibitem[Yang et~al\mbox{.}(2024b)]%
        {yang2024qwen2}
\bibfield{author}{\bibinfo{person}{An Yang}, \bibinfo{person}{Baosong Yang}, \bibinfo{person}{Beichen Zhang}, \bibinfo{person}{Binyuan Hui}, \bibinfo{person}{Bo Zheng}, \bibinfo{person}{Bowen Yu}, \bibinfo{person}{Chengyuan Li}, \bibinfo{person}{Dayiheng Liu}, \bibinfo{person}{Fei Huang}, \bibinfo{person}{Haoran Wei}, {et~al\mbox{.}}} \bibinfo{year}{2024}\natexlab{b}.
\newblock \showarticletitle{Qwen2. 5 technical report}.
\newblock \bibinfo{journal}{\emph{arXiv preprint arXiv:2412.15115}} (\bibinfo{year}{2024}).
\newblock


\bibitem[Yang et~al\mbox{.}(2024a)]%
        {SIGIR_pseudolikelihood}
\bibfield{author}{\bibinfo{person}{Shenghao Yang}, \bibinfo{person}{Weizhi Ma}, \bibinfo{person}{Peijie Sun}, \bibinfo{person}{Qingyao Ai}, \bibinfo{person}{Yiqun Liu}, \bibinfo{person}{Mingchen Cai}, {and} \bibinfo{person}{Min Zhang}.} \bibinfo{year}{2024}\natexlab{a}.
\newblock \showarticletitle{Sequential Recommendation with Latent Relations based on Large Language Model}. In \bibinfo{booktitle}{\emph{Proceedings of the 47th International ACM SIGIR Conference on Research and Development in Information Retrieval}}. \bibinfo{publisher}{Association for Computing Machinery}, \bibinfo{pages}{335–344}.
\newblock


\bibitem[Yao et~al\mbox{.}(2023)]%
        {yao2023react}
\bibfield{author}{\bibinfo{person}{Shunyu Yao}, \bibinfo{person}{Jeffrey Zhao}, \bibinfo{person}{Dian Yu}, \bibinfo{person}{Nan Du}, \bibinfo{person}{Izhak Shafran}, \bibinfo{person}{Karthik Narasimhan}, {and} \bibinfo{person}{Yuan Cao}.} \bibinfo{year}{2023}\natexlab{}.
\newblock \showarticletitle{React: Synergizing reasoning and acting in language models}. In \bibinfo{booktitle}{\emph{International Conference on Learning Representations (ICLR)}}.
\newblock


\bibitem[Zavitsanos et~al\mbox{.}(2025)]%
        {ACM_ML_FinFraud}
\bibfield{author}{\bibinfo{person}{Elias Zavitsanos}, \bibinfo{person}{Eirini Spyropoulou}, \bibinfo{person}{George Giannakopoulos}, {and} \bibinfo{person}{Georgios Paliouras}.} \bibinfo{year}{2025}\natexlab{}.
\newblock \showarticletitle{Machine Learning for Identifying Risk in Financial Statements: A Survey}.
\newblock \bibinfo{journal}{\emph{ACM Comput. Surv.}} \bibinfo{volume}{57}, \bibinfo{number}{9} (\bibinfo{year}{2025}).
\newblock


\bibitem[Zhang et~al\mbox{.}(2023)]%
        {zhang2023survey}
\bibfield{author}{\bibinfo{person}{Lingxi Zhang}, \bibinfo{person}{Jing Zhang}, \bibinfo{person}{Xirui Ke}, \bibinfo{person}{Haoyang Li}, \bibinfo{person}{Xinmei Huang}, \bibinfo{person}{Zhonghui Shao}, \bibinfo{person}{Shulin Cao}, {and} \bibinfo{person}{Xin Lv}.} \bibinfo{year}{2023}\natexlab{}.
\newblock \showarticletitle{A survey on complex factual question answering}.
\newblock \bibinfo{journal}{\emph{AI Open}}  \bibinfo{volume}{4} (\bibinfo{year}{2023}), \bibinfo{pages}{1--12}.
\newblock


\bibitem[Zhang et~al\mbox{.}(2024)]%
        {NEURIPS2024_CoA}
\bibfield{author}{\bibinfo{person}{Yusen Zhang}, \bibinfo{person}{Ruoxi Sun}, \bibinfo{person}{Yanfei Chen}, \bibinfo{person}{Tomas Pfister}, \bibinfo{person}{Rui Zhang}, {and} \bibinfo{person}{Sercan~\"{O}. Ar\i~k}.} \bibinfo{year}{2024}\natexlab{}.
\newblock \showarticletitle{Chain of Agents: Large Language Models Collaborating on Long-Context Tasks}. In \bibinfo{booktitle}{\emph{Advances in Neural Information Processing Systems}}, Vol.~\bibinfo{volume}{37}. \bibinfo{publisher}{Curran Associates, Inc.}, \bibinfo{pages}{132208--132237}.
\newblock


\bibitem[Zhao et~al\mbox{.}(2024c)]%
        {zhao-etal-2024-longagent}
\bibfield{author}{\bibinfo{person}{Jun Zhao}, \bibinfo{person}{Can Zu}, \bibinfo{person}{Xu Hao}, \bibinfo{person}{Yi Lu}, \bibinfo{person}{Wei He}, \bibinfo{person}{Yiwen Ding}, \bibinfo{person}{Tao Gui}, \bibinfo{person}{Qi Zhang}, {and} \bibinfo{person}{Xuanjing Huang}.} \bibinfo{year}{2024}\natexlab{c}.
\newblock \showarticletitle{{LONGAGENT}: Achieving Question Answering for 128k-Token-Long Documents through Multi-Agent Collaboration}. In \bibinfo{booktitle}{\emph{Proceedings of the 2024 Conference on Empirical Methods in Natural Language Processing}}. \bibinfo{publisher}{Association for Computational Linguistics}, \bibinfo{pages}{16310--16324}.
\newblock


\bibitem[Zhao et~al\mbox{.}(2024b)]%
        {zhao-etal-2024-longrag}
\bibfield{author}{\bibinfo{person}{Qingfei Zhao}, \bibinfo{person}{Ruobing Wang}, \bibinfo{person}{Yukuo Cen}, \bibinfo{person}{Daren Zha}, \bibinfo{person}{Shicheng Tan}, \bibinfo{person}{Yuxiao Dong}, {and} \bibinfo{person}{Jie Tang}.} \bibinfo{year}{2024}\natexlab{b}.
\newblock \showarticletitle{{L}ong{RAG}: A Dual-Perspective Retrieval-Augmented Generation Paradigm for Long-Context Question Answering}. In \bibinfo{booktitle}{\emph{Proceedings of the 2024 Conference on Empirical Methods in Natural Language Processing}}. \bibinfo{publisher}{Association for Computational Linguistics}, \bibinfo{pages}{22600--22632}.
\newblock


\bibitem[Zhao et~al\mbox{.}(2023)]%
        {zhao2023survey}
\bibfield{author}{\bibinfo{person}{Wayne~Xin Zhao}, \bibinfo{person}{Kun Zhou}, \bibinfo{person}{Junyi Li}, \bibinfo{person}{Tianyi Tang}, \bibinfo{person}{Xiaolei Wang}, \bibinfo{person}{Yupeng Hou}, \bibinfo{person}{Yingqian Min}, \bibinfo{person}{Beichen Zhang}, \bibinfo{person}{Junjie Zhang}, \bibinfo{person}{Zican Dong}, {et~al\mbox{.}}} \bibinfo{year}{2023}\natexlab{}.
\newblock \showarticletitle{A survey of large language models}.
\newblock \bibinfo{journal}{\emph{arXiv preprint arXiv:2303.18223}} \bibinfo{volume}{1}, \bibinfo{number}{2} (\bibinfo{year}{2023}).
\newblock


\bibitem[Zhao et~al\mbox{.}(2024a)]%
        {zhao2023financemath}
\bibfield{author}{\bibinfo{person}{Yilun Zhao}, \bibinfo{person}{Hongjun Liu}, \bibinfo{person}{Yitao Long}, \bibinfo{person}{Rui Zhang}, \bibinfo{person}{Chen Zhao}, {and} \bibinfo{person}{Arman Cohan}.} \bibinfo{year}{2024}\natexlab{a}.
\newblock \showarticletitle{FinanceMATH: Knowledge-Intensive Math Reasoning in Finance Domains}. In \bibinfo{booktitle}{\emph{Proceedings of the 62nd Annual Meeting of the Association for Computational Linguistics (Volume 1: Long Papers)}}. \bibinfo{publisher}{Association for Computational Linguistics}.
\newblock


\bibitem[Zhu et~al\mbox{.}(2021)]%
        {zhu2021tat}
\bibfield{author}{\bibinfo{person}{Fengbin Zhu}, \bibinfo{person}{Wenqiang Lei}, \bibinfo{person}{Youcheng Huang}, \bibinfo{person}{Chao Wang}, \bibinfo{person}{Shuo Zhang}, \bibinfo{person}{Jiancheng Lv}, \bibinfo{person}{Fuli Feng}, {and} \bibinfo{person}{Tat-Seng Chua}.} \bibinfo{year}{2021}\natexlab{}.
\newblock \showarticletitle{{TAT}-{QA}: A Question Answering Benchmark on a Hybrid of Tabular and Textual Content in Finance}. In \bibinfo{booktitle}{\emph{Proceedings of the 59th Annual Meeting of the Association for Computational Linguistics and the 11th International Joint Conference on Natural Language Processing (Volume 1: Long Papers)}}. \bibinfo{publisher}{Association for Computational Linguistics}.
\newblock


\bibitem[Zhu et~al\mbox{.}(2025)]%
        {zhu2025dianjin}
\bibfield{author}{\bibinfo{person}{Jie Zhu}, \bibinfo{person}{Qian Chen}, \bibinfo{person}{Huaixia Dou}, \bibinfo{person}{Junhui Li}, \bibinfo{person}{Lifan Guo}, \bibinfo{person}{Feng Chen}, {and} \bibinfo{person}{Chi Zhang}.} \bibinfo{year}{2025}\natexlab{}.
\newblock \showarticletitle{DianJin-R1: Evaluating and Enhancing Financial Reasoning in Large Language Models}.
\newblock \bibinfo{journal}{\emph{arXiv preprint arXiv:2504.15716}} (\bibinfo{year}{2025}).
\newblock


\end{thebibliography}
%%
%% If your work has an appendix, this is the place to put it.
% \appendix

\end{document}